\documentclass[runningheads]{llncs}

\usepackage{eccv}

\usepackage{eccvabbrv}

\usepackage{graphicx}
\usepackage{booktabs}

\usepackage[accsupp]{axessibility}  

\usepackage{hyperref}

\usepackage{orcidlink}

\usepackage{multirow}

\begin{document}

\title{Generating 3D House Wireframes with Semantics} 

\titlerunning{3D House WireFrames}

\author{Xueqi Ma\orcidlink{0009-0004-0203-8501} \and
Yilin Liu\orcidlink{0000-0001-7336-1956} \and
Wenjun Zhou\orcidlink{0000-0003-1790-4201} \and
Ruowei Wang\orcidlink{0009-0003-9112-1712} \and
Hui Huang\thanks{Corresponding author}\orcidlink{0000-0003-3212-0544}}

\authorrunning{X.~Ma, Y.~Liu, W.~Zhou, R.~Wang, and H.~Huang}

\institute{Visual Computing Research Center, Shenzhen University \\
\email{hhzhiyan@gmail.com}\\
}

\maketitle

\begin{abstract}

We present a new approach for generating 3D house wireframes with semantic enrichment using an autoregressive model. Unlike conventional generative models that independently process vertices, edges, and faces, our approach employs a unified wire-based representation for improved coherence in learning 3D wireframe structures. By re-ordering wire sequences based on semantic meanings, we facilitate seamless semantic integration during sequence generation. Our two-phase technique merges a graph-based autoencoder with a transformer-based decoder to learn latent geometric tokens and generate semantic-aware wireframes. Through iterative prediction and decoding during inference, our model produces detailed wireframes that can be easily segmented into distinct components, such as walls, roofs, and rooms, reflecting the semantic essence of the shape. Empirical results on a comprehensive house dataset validate the superior accuracy, novelty, and semantic fidelity of our model compared to existing generative models.
More results and details can be found on \href{https://vcc.tech/research/2024/3DWire}{https://vcc.tech/research/2024/3DWire}.

  \keywords{Wireframe \and 3D Generation \and Autoregressive Model}
\end{abstract}
\section{Introduction}
\label{sec:intro}
\begin{figure}[t]
    \centering
    \includegraphics[width=0.99\textwidth]{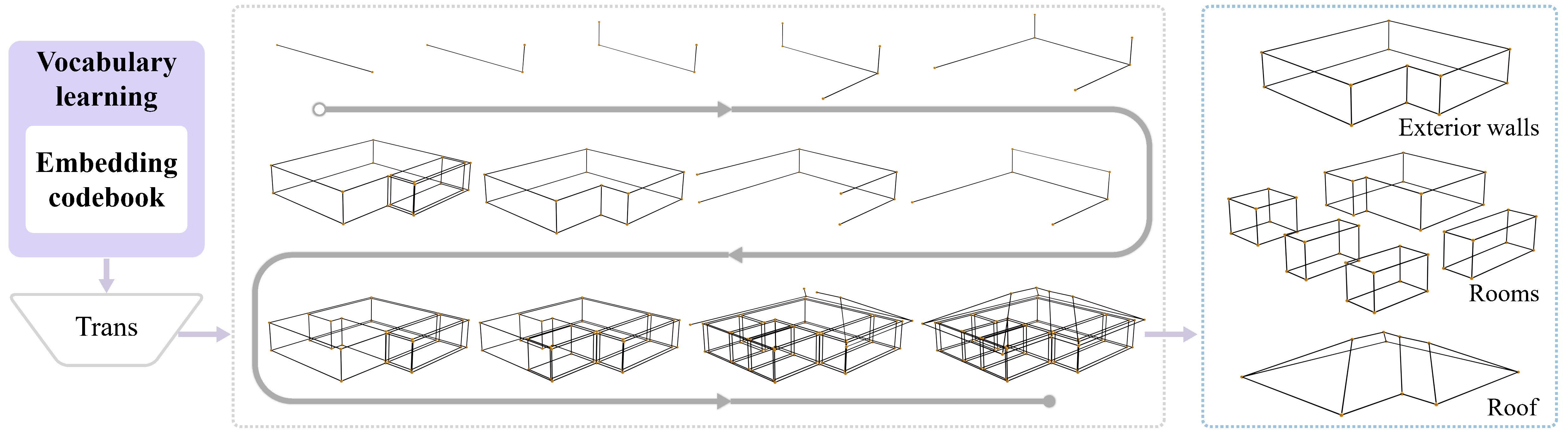}
    \caption{
        Our method creates 3D wireframes through autoregressive sampling from a trained transformer model, generating tokens from a learned geometric vocabulary. 
        These tokens are decoded into line segments to form the final wireframe. 
        Based on the nodes' connectivity, the resulting wireframes can be easily split into multiple parts, such as walls, roofs, and rooms, reflecting the underlying semantic meaning of shapes.
    }
    \label{fig:teaser}
\end{figure}

    The 3D wireframe serves as a crucial data structure in the fields of computer vision and graphics, offering a clean and compact abstraction of an object's shape through a composition of points and lines.
    This form of representation is pivotal for the creation of complex and detailed models, with a particular advantage in representing polyhedral objects such as mechanical parts, furniture, and buildings, due to its simplicity and efficacy in preserving sharp features.
    Despite its wide use and benefits, the generation of 3D wireframes remains a sophisticated and demanding process. 
    It involves the meticulous abstraction of the object's geometry into a series of precise line segments, a task that necessitates both accuracy and creativity. 
    Consequently, there is a significant need for an automated approach to efficiently generate accurate and innovative 3D wireframes.

    Early studies in wireframe generation have mainly focused on reconstructing wireframes from various data sources, like images~\cite{Yichao19Manhattan,Yichao19EndtoEnd,Pautrat23DeepLSD} or point clouds~\cite{Yujia21PC2WF,Cao23WireframeNet}. 
    These approaches typically employ learning-based techniques for detecting corners and edges, followed by a handcraft assembling process to extract the wireframe structures from the input data. 
    Despite their effectiveness in achieving high levels of accuracy, these methods do not possess the functionality to create novel wireframe data.
    In contrast, the latest research has shown an increasing interest in the learning-based 3D generation~\cite{nash2020polygen,solidgen23,siddiqui2023meshgpt}. 
    These approaches aim to learn the distribution patterns of primitive sequences (vertices, edges, faces) during the training phase. 
    Consequently, they can generate new data by sampling from these distributions and predicting primitives in an autoregressive manner.
    However, these methods treat different types of primitives (vertices, edges, faces) as distinct sequences and model them separately.
    The error in the generation of one type of primitive can propagate to the subsequent generation of other primitives, leading to a more complex learning process.
    Moreover, unlike text generation tasks, which inherently follow a sequential order, geometric primitives do not have a natural order. 
    Current methods predominantly organize these primitive sequences based on their spatial coordinates.
    The absence of high-level correlations among primitives might introduce ambiguity during the distribution modeling, which can adversely impact the quality of the generated models.
    
    In this paper, we introduce a new method for the generation of semantically grouped 3D house wireframes.
    Our method adopts a wire-based representation, which, unlike traditional methods that model vertices and edges \textit{separately}, focuses on constructing \textit{pure wire} sequences based on their semantic correlations. 
    A wireframe under our model is conceptualized as a graph where nodes correspond to wires (line segments), and edges denote the connectivity among them. 
    Notably, a wireframe may comprise disconnected components—for example, the roof, or the exterior walls of a house are usually not connected to the interior rooms. 
    We model each disconnected component as a distinct sub-graph, organizing wires based on their topological connectivity to mirror the semantic relationships among them. 
    The sequence of wires is established through a breadth-first search (BFS) traversal of the graph, ensuring the generation of a wireframe that is both coherent and semantically structured, as shown in Fig.~\ref{fig:teaser}.
    
    The wireframe generation is split into two stages.
    In the initial stage, we learn a vocabulary of latent geometric tokens representing the wireframe's wires.
    This involves employing a Graph Convolutional Network (GCN)~\cite{zhang2019graph} to encode the local geometric and topological features of the wires, complemented by an attention-based encoder designed to extract the global information of the wireframe.
    The subsequent stage utilizes a transformer-based~\cite{Vaswani17transformer} decoder to autoregressively produce a sequence of tokens from the trained vocabulary.
    These tokens are then decoded into spatial coordinates to construct the wires, which are used in the assembly of the final wireframe. Extensive experiments show that our method can faithfully generate 3D house wireframes with semantics, which outperforms state-of-the-art generative models with higher accuracy and novelty, while enhancing the semantic integrity of the generated 3D wireframes.

    In summary, our approach introduces two significant contributions:
    
    • We propose a wire-based representation model for learning wireframe distributions, significantly enhancing the accuracy of 3D wireframe generation.
    
    • Our method introduces a semantic-aware sequence construction technique that reduces ambiguity in the learning phase. 
    Moreover, it allows for the wireframe to be segmented into distinct parts, each reflecting the semantic essence of the underlying 3D house model.

\section{Related Work}
\label{sec:rw}

The concept of wireframes as a fundamental data structure has been extensively explored over several decades. 
This section covers three primary areas of research that are closely related to our study: wireframe reconstruction, floorplan generation, and the application of autoregression in 3D model generation.

\subsection{Wireframe reconstruction}

\paragraph{Reconstructing Wireframes from Images:} 
A significant area of research focuses on deriving wireframe representations from images of man-made environments~\cite{Kun18Learning,Yichao19EndtoEnd,Xue2019AFM,Wenchao22How3D,Xue23Holistically}. 
This involves the use of deep convolutional neural networks~\cite{zewen2022CNN} to accurately detect vertices and line segments, facilitating the creation of wireframe models. 
Various studies have introduced different geometric priors to enhance this process, including a trainable Hough transform block~\cite{Yancong2020DeepHL}, the utilization of global structural regularities~\cite{Yichao19Manhattan}, and the exploitation of deep image gradients~\cite{Pautrat23DeepLSD}. 
Additionally, recent efforts~\cite{Xue23NEAT,Yicheng22LC2WF} have extended these techniques to generate wireframes from sequences of images, demonstrating advancements in temporal data processing for 3D reconstruction.

\paragraph{Reconstructing Wireframes from Point Clouds:}
Recent research focuses on deriving 3D wireframes from point clouds. The PC2WF model~\cite{Yujia21PC2WF} encodes point clouds to feature points and connects them to establish 3D wireframes. 
WireframeNet ~\cite{Cao23WireframeNet} refines the point cloud using a medial axis transform (MAT) before predicting edge points and constructing the wireframe based on their connectivity.
Matveev et al.~\cite{Matveev21Parametric} use a scalar distance field for creating a wireframe's topological graph from detected corners and curves. 
Furthermore, Tan et al.~\cite{Xuefeng22Coarse} introduce a method that integrates coarse and fine pruning modules for wireframe optimization, employing particle swarm optimization to maintain the wireframe's correct topology.

However, current research predominantly utilizes images and point clouds as prerequisites for generating 3D wireframes, highlighting a gap in the capability to produce novel 3D wireframes without pre-specified conditions. 
This underscores the ongoing challenge of unconditional 3D wireframe generation.

\subsection{Floorplan Generation}
Wireframes serve a pivotal role in floor plan generation, where building layouts are depicted as 2D nodes and edges. 
Wu et al.~\cite{Wenming19PlanGen} pioneered house layout generation by simulating the human design process, initially determining room locations followed by wall placement. 
HouseGAN~\cite{nauata2020housegan} and House-GAN++~\cite{nauata2021houseganpp} utilize graph-constrained relational and conditional GANs~\cite{pathak2016context} to create realistic and diverse house layouts. 
HouseDiffusion~\cite{shabani2022housediffusion} adopts bubble diagrams to generate vector floor plans via a dual diffusion strategy. 
FLNet~\cite{Upadhyay22FLNet} merges graph convolutional networks with spatial layout networks for crafting floor layouts within user-defined constraints. 
Graph2Plan~\cite{Ruizhen20Graph2Plan} integrates user feedback to automate floor plan generation from building boundaries and layout diagrams. WallPlan~\cite{Jiahui22WallPlan} models wall structures as graphs, using intersections as nodes and sections as edges, to incrementally develop floor plans through graph traversal.
RoomFormer~\cite{yue2023connecting} utilizes a Transformer-based approach for generating multiple room polygons in floorplan reconstruction.

Predominantly, these methodologies concentrate on generating 2D floor plans, \textit{relying} on constraints such as graphs~\cite{nauata2020housegan, nauata2021houseganpp, shabani2022housediffusion, Ruizhen20Graph2Plan, Upadhyay22FLNet}, building boundaries~\cite{Ruizhen20Graph2Plan, Wenming19PlanGen, Upadhyay22FLNet, Shidong2023ActFloor, Jiahui22WallPlan}, textual inputs~\cite{sicong2023tell2design, qi2020Intelligent}, or point clouds~\cite{yue2023connecting}. 
Converting these 2D layouts into 3D architectural models typically necessitates additional post-processing.

\subsection{Autoregression for 3D Generation}
Recently, auto-regressive models have made significant strides in 2D image generation~\cite{esser2021taming, van2016conditional, razavi2019generating} and have also extended to 3D tasks~\cite{sun2020pointgrow, cheng2022autoregressive, nash2020polygen, yan2022shapeformer, siddiqui2023meshgpt}.
Autosdf~\cite{mittal2022autosdf} tackles multimodal 3D tasks using an autoregressive prior, grounded in a discretized, low-dimensional latent representation. 
Ibing et al.~\cite{Ibing2023Octree} use octrees for hierarchical 3D shape representation and introduce adaptive compression to address sequence length, improving autoregressive model efficiency.
PolyGen~\cite{nash2020polygen} is an autoregressive method for generating 3D meshes. 
It employs transformer~\cite{Vaswani17transformer} and pointer networks~\cite{Vinyals2015Pointer} to predict sequences of vertices and faces in the mesh. 
Similarly, MeshGPT~\cite{siddiqui2023meshgpt} learns and encodes quantized embeddings of triangular mesh geometry in order to autonomously generate highly compact triangular meshes.
SolidGen~\cite{solidgen23} utilizes transformers and pointer networks to directly model B-rep and predict vertices, edges, and faces in an autoregressive manner. 

While PolyGen~\cite{nash2020polygen} and SolidGen~\cite{solidgen23} share similarities with our approach, they serialize primitives (vertices, edges, faces) separately. 
This independent processing can accumulate errors; inaccuracies in vertices may impact edges and faces. 
Moreover, these methods construct primitive sequences based on coordinates, neglecting semantic relationships. 
In contrast, our method emphasizes exploiting these semantic connections to sequence line segments, creating coherent wireframes that are semantically rich and hierarchically organized.
\section{Methodology}
\label{sec:method}

In 3D design, professional designers construct wireframes by drawing each line segment individually. 
Similarly, large language models use a sequential approach to build complex linguistic structures. 
Our methodology synthesizes these principles, adopting a line-by-line generation approach to craft 3D wireframes.
As shown in Fig.~\ref{fig:teaser}, we first learn a geometric embedding vocabulary from a large collection of 3D wireframes.
Each line segment is quantized into a latent space, facilitating its encoding and subsequent processing as detailed in Sec.~\ref{sec:method-quantization}.
Following this, we employ a transformer-based model trained to predict the subsequent code within our predefined vocabulary, gradually generating the corresponding 3D wireframes~\ref{sec:method-transformer}.

\subsection{Learning Quantized Line Segment Embeddings}
\label{sec:method-quantization}
As demonstrated by SolidGen~\cite{solidgen23} and MeshGPT~\cite{siddiqui2023meshgpt}, the development of a quantized feature or embedding vocabulary is foundational for effective 3D shape generation.
Our approach begins with the derivation of a latent code for each line segment through an encoder, denoted as $\mathrm{E}$, followed by the quantization of these codes into a geometric codebook via Residual Quantization~\cite{lee2022autoregressive}.
We define a 3D wireframe \( \mathcal{W} \) as a set of line segments:
\begin{equation}
    \mathcal{W} = (l_1, l_2, l_3, \dots, l_N),
\end{equation}
where \( \mathcal{W}\) consists of \(N\) line segments. Each segment \( l_i \in \mathbb{R}^{n_\mathrm{in}} \) is characterized by:
    1) Coordinates of its endpoints,
    2) Length of the segment
    3) Directional orientation
    4) Angle between it and adjacent segments
    5) Coordinates of its midpoint. 
All these features are quantized to an integer of range [0,128) and then embedded into a 196-dimensional vector.

\begin{figure}[t]
    \centering
    \includegraphics[width=0.99\textwidth]{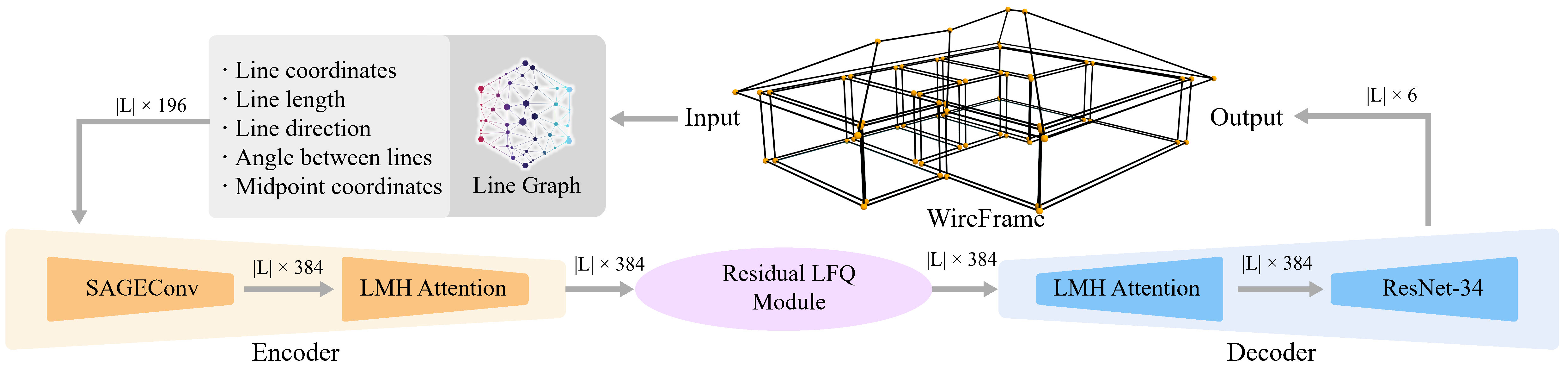}
    \caption{Pipeline of learning the geometric vocabulary of line segments.}
    \label{fig:ae-pipeline}
\end{figure}

As shown in Fig.~\ref{fig:ae-pipeline}, to enhance neighborhood data representation and simplify the geometric vocabulary, we first transform the feature of each line segment \( l_i \) into a \( n_\mathrm{z} \)-dimensional vector \( z_i \) using an encoder $\mathrm{E}(\mathcal{W})$, $z=384$ in all our experiments.
The encoder consists of a graph convolutional encoder $\mathrm{E_G}(\mathcal{W})$ and an attention-based information exchanger $\mathrm{E_A}(\mathcal{W})$. 
The encoder $\mathrm{E_G}: (\mathbb{R}^{N \times n_{in}}) \rightarrow \mathbb{R}^{N \times n_z} $ treat line segments as nodes, with their adjacency defining the graph edges.
We utilize SAGEConv layers \cite{Hamilton17SAGEConv} to project node features into latent space, capturing the local geometric feature of each segment.

Given that the input wireframe $\mathcal{W}$ may contain disconnected sub-graphs, where two nodes that are spatially close might not be topologically connected, relying solely on graph convolutions might be insufficient for information exchange.

To mitigate this, we introduce an attention-based exchanger $\mathrm{E_A}: (\mathbb{R}^{N \times n_z}) \rightarrow \mathbb{R}^{N \times n_z} $, specifically a Local Multi Head Attention(LMH Attention) \cite{Aurko21Efficient, beltagy2020longformer}, comprised of multiple self-attention layers.
It allows for a complete information exchange across the entire structure, ensuring the extracted features are not only topologically informed but also contextually enriched.

Based on the high-dimensional feature of each segment $l_i$, we then learn a geometric embedding codebook through Residual Quantization (RQ).
Similar to MeshGPT~\cite{siddiqui2023meshgpt}, we learn the quantized code for each vertex instead of each line segment to get better generalizability.
This involves assigning the segment features to their endpoints and normalizing by the count of adjacent segments, effectively distributing the feature information across the vertices.
For a given codebook \( C \), a Residual Quantization (RQ) with \( D \) layers of depth can represent a feature \( z \) as follows:

\begin{equation}
   (m_1, m_2, \ldots , m_D) \in [M]^D = \mathcal{RQ}(z; C, D),
\end{equation}
where, \(m_d\) is the code of \(z\) at depth \(d\).

This quantization process involves comparing each vertex feature, against all codes in the codebook, thus increasing the model's computational complexity. 
To overcome this challenge, we employed the lookup-free quantization (LFQ)~\cite{yu2023language}. 
LFQ quantizes vertex features by treating them as a Cartesian product of single-dimensional variables, which eliminates the need for the codebook lookup step typically required in traditional quantization. 

After the quantization, we stack the quantized vertex features according to the indices of the line segment vertices to form the features of the line segment. 
These features are then decoded into six distinct coordinate values via a decoder network comprising LMH Attention layers and a 1D ResNet34 \cite{Kaiming16ResNet} architecture. 
We guide the reconstruction process of 3D wireframes by applying cross-entropy loss to discrete 3D wireframe coordinates, coupled with a commitment loss for vertex embeddings. 
Additionally, we employ an entropy penalty~\cite{yu2023language} to enhance the utilization of the codebook. 
This strategy not only boosts the network's confidence in its predictions but also promotes a more diverse usage of the codebook entries. 
More details can be found in supplementary. 
Once we have trained the encoder and the codebook, we can use the quantized feature of each line segment as a token sequence to further train the autoregressive model.

\begin{figure}[t]
    \centering
    \includegraphics[width=\textwidth]{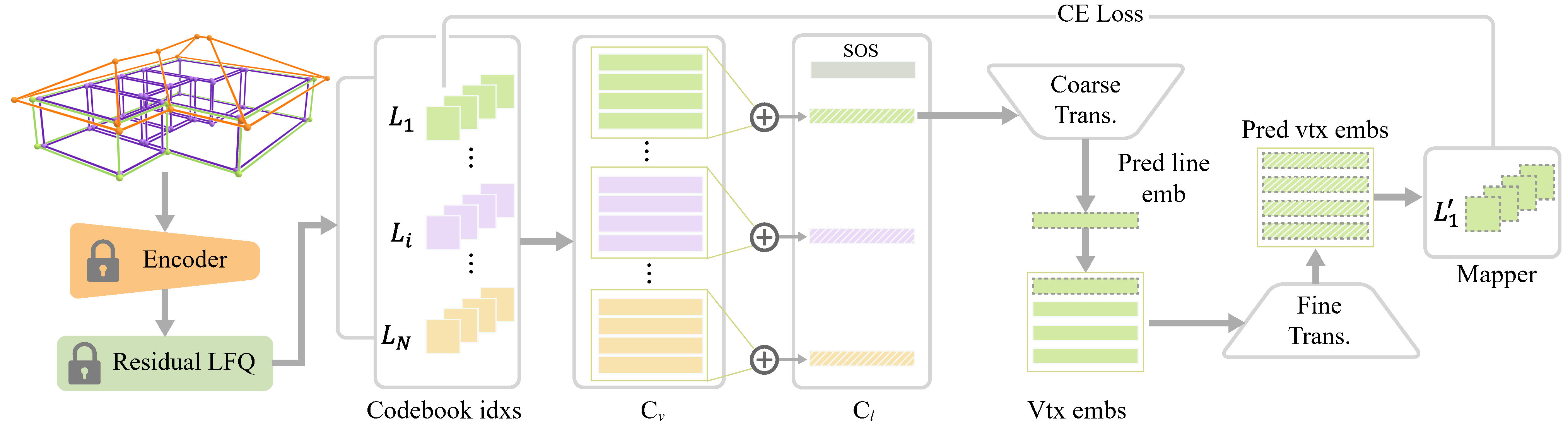}
    \caption{
    Pipeline of the transformer training. 
    Firstly, the wireframe is encoded by the encoder to extract features and undergo Residual LFQ \cite{yu2023language} to obtain code indices of each line segment. 
    These indices are then split and transformed into code embeddings for vertices ($C_v$) and are recovered to the code embeddings of line segments ($C_l$). 
    The line embeddings are progressively predicted (Pred line emb) by the coarse transformer and refined to the vertex embedding (Pred vtx embs) by the fine transformer.
    They are finally transformed back to codebook indices through a mapper, and these indices are then optimized using a loss function to generate high-quality wireframes.
    }
    \label{fig:trans_pipeline}
\end{figure}

\subsection{3D Wireframe Generation with Transformers}
\label{sec:method-transformer}
Leveraging the quantized features of each line segment, we employ a transformer model trained to predict the next segment in the sequence autoregressively. 
Autoregressive models, by design, generate sequences where each subsequent token depends on the previous ones. 
In the context of 3D wireframe generation with transformers, establishing a predetermined sequence for line segment generation is crucial. 
This sequence not only ensures the structural integrity of the wireframes but also aids the model in learning the intricate geometric shapes.

Unlike previous methods~\cite{siddiqui2023meshgpt, solidgen23} that rely solely on coordinate-based token sequences, our approach sequences tokens based on the semantic meaning of the line segments.
Adapting PolyGen's~\cite{nash2020polygen} strategy, we organize wireframe junctions and segments in a z-y-x hierarchical order.
Then we treat the line segments of the wireframe as graph nodes and the junctions as graph edges, thereby transforming the 3D wireframe into a graph structure. 
Note that house wireframes often comprise multiple disconnected subgraphs.
Thus, a Breadth-first search algorithm is applied to group segments in the same subgraph, thereby ensuring sequential generation of segments from the same object.

As shown in Fig.~\ref{fig:trans_pipeline}, our model utilizes a decoder-only transformer architecture to predict segment sequence indices derived from a previously established codebook (see Section~\ref{sec:method-quantization}).
In this architecture, indices are transformed into learnable embeddings, complemented by three distinct types of encoding:
1) a learnable discrete positional encoding, which marks the position of each code in the code sequence. 
2) a vertex positional encoding, indicating the position of vertices within a line segment. 
3) a quantizer level encoding to differentiate between various depths of residual quantization levels. 
This results in encoded vertex code sequences  \( C_v \), with a length of \( |C_v| = 2 \cdot D \cdot N \).

To effectively learn the discrete code stack with depth $D$ extracted by Residual LFQ, we trained the transformer in two stages: coarse transformer and fine transformer~\cite{lee2022autoregressive}. 
During the coarse transformer stage, \( C_v \) is reshaped into a \( 2 \cdot D \times N \). 
Subsequently, by merging these \( 2 \cdot D \) dimensions, we acquire the line code sequences \( C_l \), whose length is \( |C_l| = N \). Then, we utilize the coarse transformer to predict the line embeddings autoregressively.
During the fine transformer stage, we predict each line segment's vertex embeddings along the depth dimension based on each predicted line embeddings.

During the inference time, the transformer can generate a token sequence autoregressively.
The sequence's corresponding embeddings, fetched from the codebook, are decoded to form the 3D wireframe. 
Duplicate vertices are merged as a final step to construct the wireframe. 

\subsection{Implementation Details}
During the learning process of the line segment vocabulary, our residual quantization use a depth of \( D = 2 \), representing each line segment by \(2 \cdot D=4\) embeddings. 
Our codebook size is 8192. 
Since we use Residual LFQ, each embedding vector has a dimension of \(\log_2^{8192}=13\). 
We employ two distinct, non-shared codebooks for richer information learning. 
Subsequently, the decoder predicts the coordinates of line segments, distributed across 128 categories, achieving spatial discretization with a resolution of \(128^3\). 
This encoder-decoder network model was trained on 8 RTX 3090 GPUs for approximately 2 days.
For the transformer, we employed a 12+2 layers coarse-to-fine decoder-only transformer model. The model was trained for about 5 days on 8 RTX 3090 GPUs.

Both the autoencoder and transformer were written in PyTorch and trained using the ADAM optimizer, with a learning rate of \( 1 \times 10^{-4} \). For the transformer, the learning rate linearly increases to a maximum of \(lr_{\max}=1 \times 10^{-4} \) during the initial \(t_0=10\) epochs. It then gradually decreases following a cosine decay schedule \(lr_{\max}*0.5*{(1+\cos(\frac{t-t_0}{T-t_0}\pi))}\) until it reaches a minimum of \( 1 \times 10^{-6} \).
\section{Experiments}

We compare our method with the state-of-the-art generative model~\cite{nash2020polygen,siddiqui2023meshgpt} to evaluate the performance of our method. 
To the best of our knowledge, we are the first to propose a method for generating 3D house wireframes with semantic enrichment.
While PolyGen \cite{nash2020polygen} and MeshGPT \cite{siddiqui2023meshgpt} aim for generating 3D models, they do not explicitly construct the wireframe structure.
We modified PolyGen and MeshGPT to shift their focus from predicting triangle or polygon vertices to line segments, allowing for a fair comparison with our method. 
We also conduct a series of ablation studies to verify the effectiveness of our technical contributions and provide an in-depth analysis of our approach.

\subsection{Dataset and Metrics}
\label{sec:data}

Since there is no existing dataset for 3D wireframe generation, we created a new dataset for this task. We first extracted corners from the house layouts in the RPLAN dataset \cite{Wenming19PlanGen} to obtain the basic wireframes of houses. Then we used the straight skeleton \cite{aichholzer1996novel,aichholzer1996straight} to construct wireframes of the house roofs. This process resulted in a 3D house wireframe dataset with approximately 78,000 wireframes. Various data augmentation techniques, including rotation, random shifts, and scaling, were employed to facilitate training. To ensure the wireframe fit within the transformer's context window, we selected only wireframes with fewer than 400 line segments for our training set. Detailed information on the augmentation processes and data splits is provided in the supplementary.

Evaluating 3D wireframe generation is challenging due to the absence of ground truth. We adopted established metrics from previous works \cite{luo2021diffusion, zeng2022lion, Zhou2021PVD}, including Minimum Matching Distance (MMD), Coverage (COV), and 1-Nearest Neighbor (1-NN), based on Chamfer Distance (CD) and Earth Mover’s Distance (EMD). To assess the structural integrity of the generated wireframe, we also examined vertex-line segment connectivity with Two-Line-Connected Vertex Proportion (2L-CVP) and Three-Line-Connected Vertex Proportion (3L-CVP). 
These metrics measure the proportion of vertices connected to at least two or three line segments, respectively, indicating wireframe structural validity. 
Additionally, we use KL divergence (KLD) to compare the distributions of the number of connected components in the generated samples with the test data.

\subsection{Quantitative Evaluations}

As shown in \cref{tab:quantitative}, our method surpasses the comparative approaches across all evaluation metrics. 
Regarding COV, our method achieves high scores in both the CD-based and EMD-based measures, indicating a strong spatial correspondence between our wireframes and the test set. 
In terms of MMD, the differences between our wireframes and the test set are minimized. 
The results of 1-NN further validate the superiority of our method, as it maintains a high level of consistency with the test set wireframes. 
As for Structural Validity, our wireframes exhibit high coherence while preserving a spatial distribution highly consistent with the test set. 
With respect to KLD, our results more closely align with the real data distribution compared to other methods.
These results confirm the effectiveness and superiority of our method in generating 3D house wireframes.

To further evaluate the quality of the generated 3D house wireframes, we conducted a user study comparing wireframes produced by different methods. Participants used an intuitive interface to view pairwise 3D wireframe models and selected the best-quality wireframe based on topological rationality. 60 participants each made 24 sets of choices.

As shown in \cref{tab:user-study}, participants showed a preference for the wireframes generated by our method compared to other methods. 
Specifically, our method received a quality rating of 0.84 compared to PolyGen and 0.75 compared to MeshGPT. 
These ratings were based on the frequency with which participants ranked our method as higher quality than the other methods, accounting for 92\% and 87\% respectively. 
Note that compared to ground truth wireframes, our method still achieved a relatively objective score of -0.13, indicating a good approximation to real wireframe models.
We also conducted a quantitative experiment on a subset of the ABC dataset~\cite{Koch_2019_CVPR}, comprising 24,634 samples of planar shapes (see \cref{tab:quantitative-abc}). Compared with the house dataset, the ABC dataset exhibited a greater diversity of shapes. As a result, the performance of all methods suffered some decline. However, our method still outperformed the others and produced the best results.
More details can be found in the supplementary. 

\begin{table}[t!]
    \captionsetup{skip=1pt}
    
    \caption{Quantitative comparison for unconditional wireframe generation on the 3D house wireframe dataset. CD is multiplied by \( 10^3 \) and EMD is multiplied by \( 10^2\).}
    \label{tab:quantitative}
    \centering
    \setlength{\tabcolsep}{5pt}
    \resizebox{\linewidth}{!}
    {
    \begin{tabular}{@{}lccccccccc@{}}
        \toprule
        \multicolumn{1}{c}{\multirow{2}{*}{Model}} & \multicolumn{2}{c}{COV (\%, \textuparrow)} & \multicolumn{2}{c}{MMD (\textdownarrow)} & \multicolumn{2}{c}{1-NN (\%)} & \multicolumn{2}{c}{Struct. Valid. (\%, \textuparrow)} & \multicolumn{1}{c}{\multirow{2}{*}{KLD}} \\
        \cmidrule(lr){2-3} \cmidrule(lr){4-5} \cmidrule(lr){6-7} \cmidrule(lr){8-9}
             & CD & EMD & CD & EMD & CD & EMD & 2L-CVP & 3L-CVP \\  
        \midrule
        PolyGen \cite{nash2020polygen}      & 38.67 & 47.95 & 8.67 & 6.43  & 74.43  & 67.65 & 81.47 & 75.80 & 12.75 \\
        MeshGPT \cite{siddiqui2023meshgpt}  & 54.78 & 54.29 & 9.13 & 6.27  & 64.61  & 61.70 & 80.91 & 70.77 & 8.98\\
        Ours                                & \textbf{56.15} & \textbf{58.64} & \textbf{8.11} & \textbf{5.75}  & \textbf{55.21}  & \textbf{51.35} & \textbf{99.53} & \textbf{99.26} & \textbf{0.73}\\
        \bottomrule
        
    \end{tabular}
    }
\end{table}

\begin{table}[t]
\centering
\caption{
    Results of the user study on evaluating wireframes generated by PolyGen, MeshGPT, our method, and ground truth.
    }
\label{tab:user-study}
\setlength{\tabcolsep}{5pt}
\begin{tabular}{@{}lcccc@{}}
\toprule
    & PolyGen & MeshGPT & Ours & Ground Truth \\ 
    \midrule
    PolyGen \cite{nash2020polygen}      & ---   & -0.67 & -0.84 & -0.93 \\
    MeshGPT \cite{siddiqui2023meshgpt}  & 0.67  & ---   & -0.75 & -0.83 \\
    Ours                                & 0.84  & 0.75  & ---   & -0.13  \\
    Ground Truth                        & 0.93  & 0.83  & 0.13   & ---  \\
    \bottomrule
\end{tabular}
\end{table}

\begin{figure}[th!]
    \centering
    \includegraphics[width=0.95\textwidth]{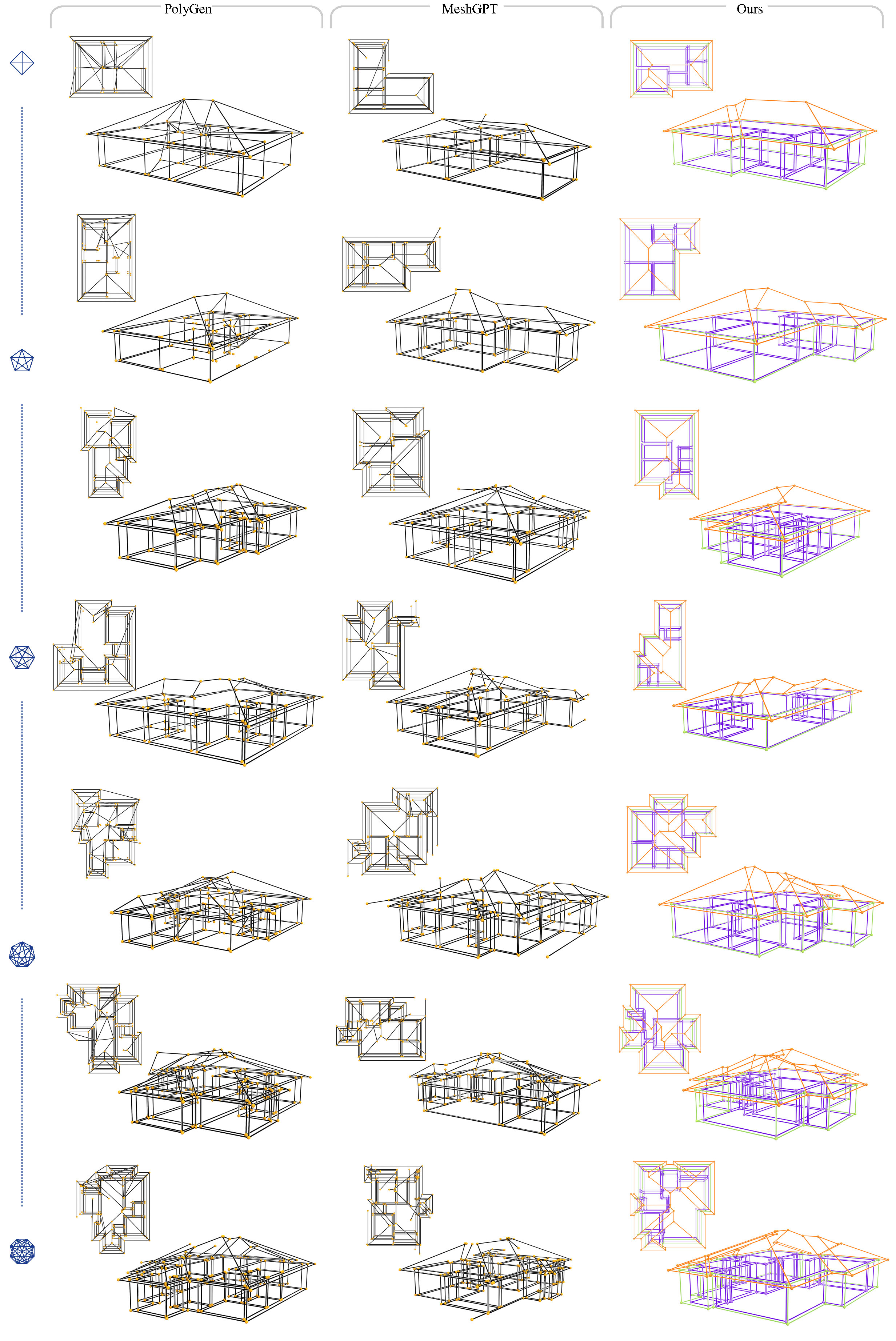}
    \caption{Qualitative comparison of 3D house wireframes. Compared to baselines, our method produces valid wireframes with high geometric fidelity and greater simplicity.}
    \label{fig:qualitative}
\end{figure}

\begin{figure}[t!]
    \centering
    \includegraphics[width=0.99\textwidth]{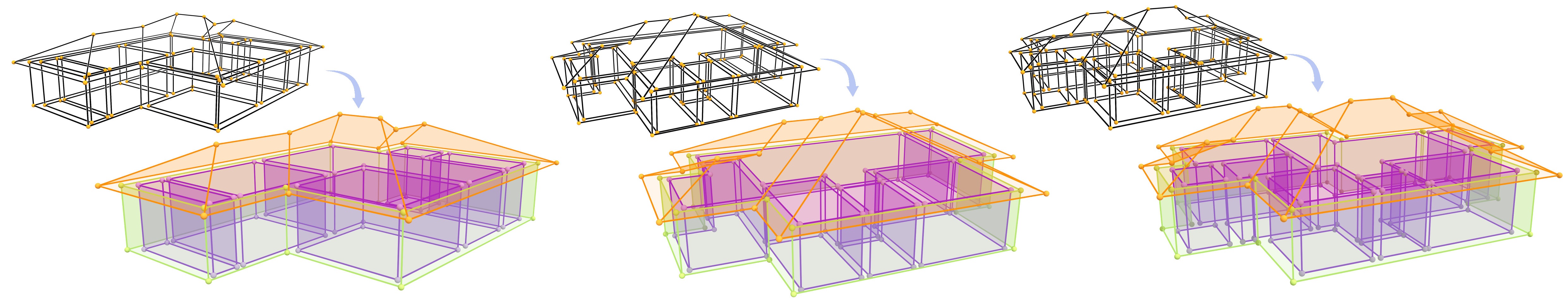}
    \captionsetup{skip=5pt}
    
    \caption{
    The resulting wireframe can be easily converted into a mesh model.
    }
    \label{fig:recon}
\end{figure}

\begin{figure}[t!]
    \centering
    \includegraphics[width=0.99\textwidth]{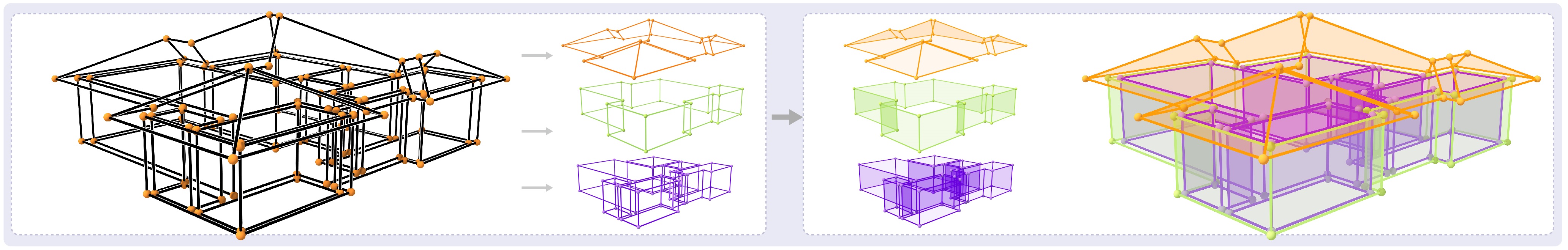}

    \caption{The reconstructed wireframe model can be easily split into several components. We also show their corresponding mesh on the right.}
    \label{fig:three-comp}
\end{figure}

\begin{figure}[t!]
    \centering
    \includegraphics[width=0.99\textwidth]{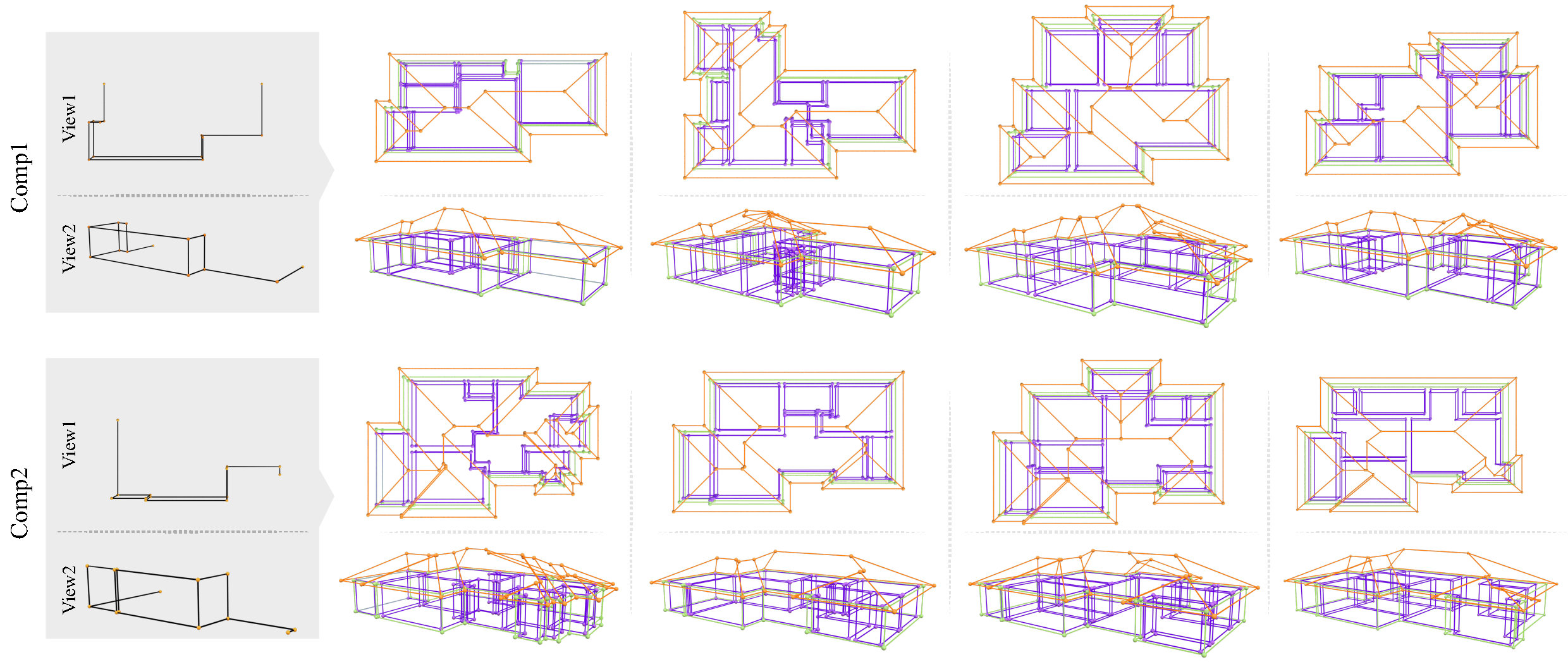}

    \caption{
        Our method has the capability to generate multiple potential completions for a given partial wireframe.
    }
    \label{fig:completion}
\end{figure}

\begin{figure}[t]
    \centering
    \includegraphics[width=0.98\linewidth]{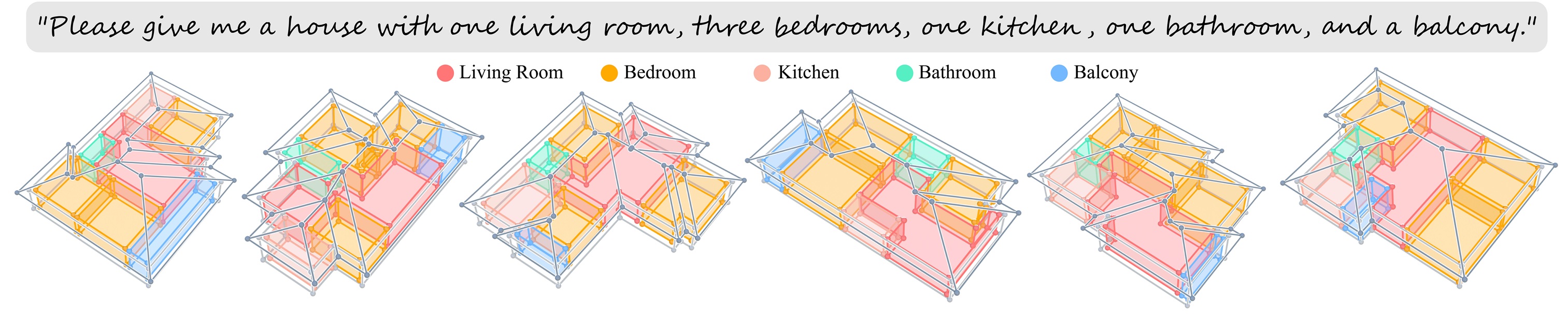}
    \caption{
        Samples of 3D house wireframes generated based on text conditions.
    }
    \label{fig:text2wf}
\end{figure}

\begin{figure}[t]
    \centering
    \includegraphics[width=0.98\linewidth]{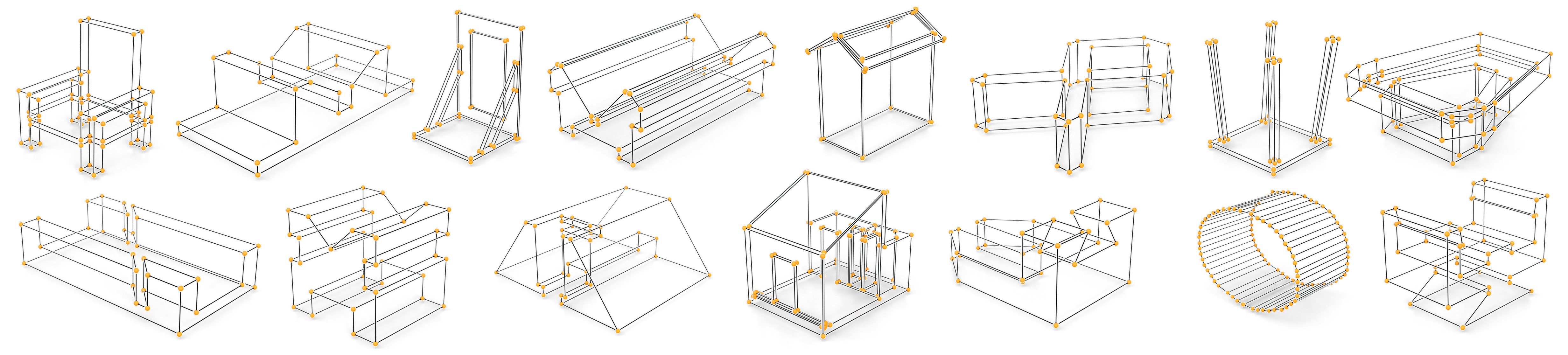}

    \caption{
        Samples generated by our method, trained on the ABC dataset~\cite{Koch_2019_CVPR}.
    }
    \label{fig:abc-gen}
\end{figure}

\begin{table}[t!]
    \caption{Quantitative comparison for unconditional generation on the ABC dataset.}
    \label{tab:quantitative-abc}
    \centering
    \setlength{\tabcolsep}{5pt}
    {
    \begin{tabular}{@{}lccc@{}}
        \toprule
        Model & COV (\%, \textuparrow) & MMD (\textdownarrow) & 1-NN (\%) \\ 
        \midrule
        PolyGen \cite{nash2020polygen}      & 39.94   & 25.53 & 75.87 \\
        MeshGPT \cite{siddiqui2023meshgpt}  & 42.38  & 24.62   & 67.65 \\
        Ours                                & \textbf{44.10}  & \textbf{22.12}  & \textbf{62.96}   \\
        \bottomrule 
        
    \end{tabular}    
    }
\end{table}

\subsection{Qualitative Evaluations}
In \cref{fig:qualitative}, we provide a qualitative comparison between 3D house wireframes generated by our approach and those from existing methods. 
PolyGen often results in floating vertices and line segments due to its sequential process of generating vertices before connections, which suffers from an error accumulation and can negatively impact the coherence of the wireframe. 
MeshGPT also shows similar issues, attributed to its neglect of semantic relationships between wireframe components like exterior walls, rooms, and roofs.
Conversely, our method excels in generating structurally sound and semantically rich wireframes, highlighting clear distinctions between exterior walls, rooms, and roofs. 
This not only improves the interpretability of the wireframes but also significantly reduces floating vertices and extraneous line segments, enhancing both accuracy and structural integrity. 
The resulting wireframe can be easily split into multiple components, such as walls, roofs, and different rooms, based on the connectivity of the line segments, as shown in \cref{fig:three-comp}.
Moreover, our wireframes are readily convertible into mesh models, as illustrated in \cref{fig:recon}.
Also, it can produce multiple possible completions for a given partial wireframe since our model generates wireframe sequences in a probabilistic manner, as shown in \cref{fig:completion}.
We included text-conditioned results in \cref{fig:text2wf}. 
Features from the input text were incorporated via cross-attention, enriching the wireframe generation process. 
\cref{fig:abc-gen} demonstrates our method's capacity to generate complex objects.

\begin{figure}[t]
    \centering
    \includegraphics[width=0.99\textwidth]{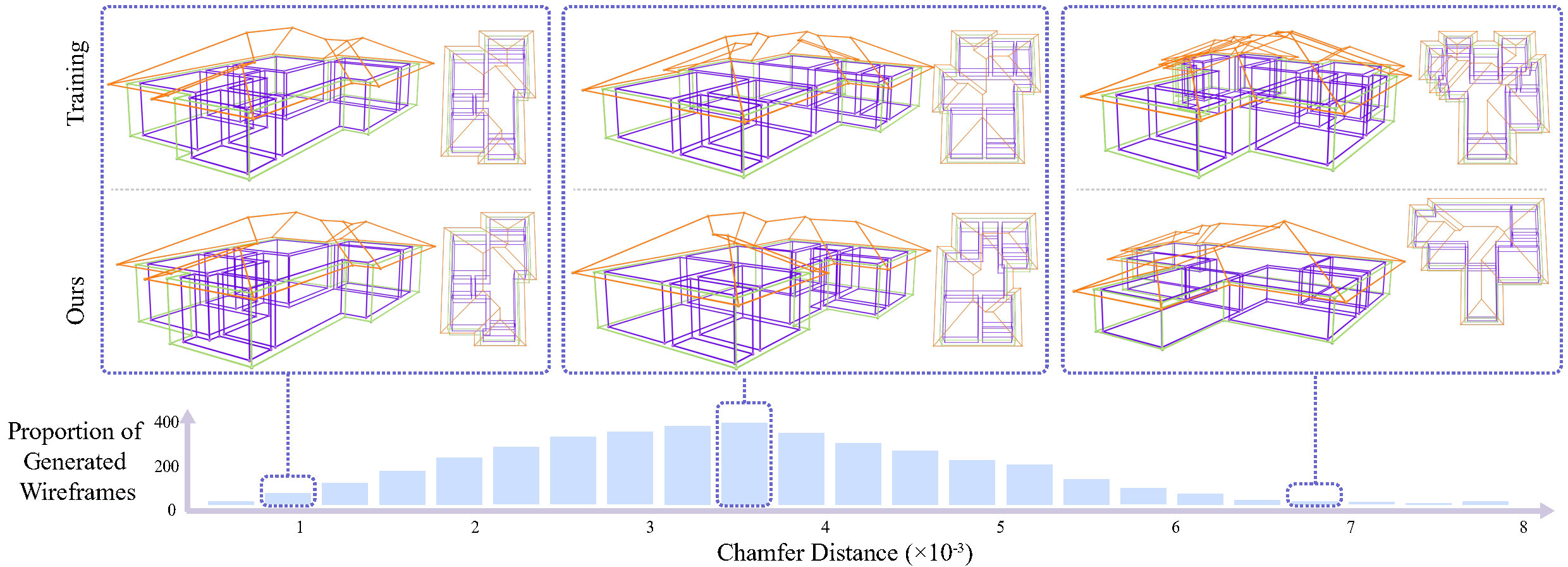}
    
    \caption{
      Wireframe novelty analysis on the 3D house wireframe dataset.
      We plot the distribution of 4096 samples generated by our method and assessed their similarity to the distribution of the training dataset based on the Chamfer Distance (CD). Our method is capable of generating wireframes that are both similar (low CD) and novel (high CD) compared to the training distribution. 
    }
    \label{fig:novelty}
\end{figure}

\subsection{Wireframe Novelty Analysis}

Inspired by previous methods \cite{Jingyu2022Wavelet, Ziya2023ICCV, siddiqui2023meshgpt}, we analyzed the novelty of the generated wireframes compared to the training dataset samples. Using our method, we generated 4096 wireframes and found the most similar samples in the training set (with the lowest Chamfer distance, CD). \cref{fig:novelty} shows a histogram of the generated wireframes. When the CD is small, our method's wireframes cover the training samples well. As CD values increase, our wireframes show more significant differences from the most similar training structures, demonstrating our method's ability to create novel wireframes. Even in the histogram's middle region, where most generated wireframes are located, our wireframes exhibit structural differences from training samples, indicating significant diversity. More details can be found in the supplementary material.

\subsection{Ablation Studies}

\begin{table}[t!]
    \caption{
        Ablations of our design choices on the 3D house wireframe dataset. The results indicate a significant decrease in performance when any of them are removed.
    }
    \label{tab:ablation}
    \centering
    \setlength{\tabcolsep}{5pt}
    \begin{tabular}{@{}lccccc@{}}
        \toprule
        Method & COV$\uparrow$ & MMD$\downarrow$ & 1-NN  & 2L-CVP$\uparrow$ & 3L-CVP$\uparrow$\\ \midrule
        w/o Encoder LMH Attention   & 49.07 & 10.98 & 68.72 & 64.97 & 59.20 \\
        w/o Residual LFQ            & 50.02 & 9.24  & 69.87 & 69.17 & 63.47 \\
        w/o Coarse-to-Fine          & 52.56 & 9.08  & 66.20 & 73.97 & 68.77 \\
        w/o Semantic Order          & 51.33 & 9.07  & 67.27 & 72.42 & 67.69 \\
        \midrule
        Ours                    & \textbf{56.15} & \textbf{8.11}  & \textbf{55.21} & \textbf{99.53} & \textbf{99.26} \\
        \bottomrule
    \end{tabular}
\end{table}

To verify our technical contributions, we conducted ablation studies on four major components of our model: 
1) The LMH attention layers in our autoencoder; 
2) The geometric codebook implemented by a Residual LFQ; 
3) A coarse-to-fine transformer when decoding the sequence; 
4) The semantic sequence reordering when training the transformer. 
As shown in \cref{tab:ablation}, removing LMH Attention layers resulted in intersecting line segments, highlighting their role in encoding spatial relationships. Replacing Residual LFQ with VQ reduced model quality due to VQ's limited effectiveness with large vocabularies. Discontinuing the coarse-to-fine strategy increased sequence length, raising training difficulty and affecting quality. Removing semantic sorting of line segments in the training data and sorting only in the z-y-x sequence also degraded quality. These findings demonstrate that each component significantly impacts overall performance. LMH Attention layers extract geometric features, while Residual LFQ, a coarse-to-fine strategy, and semantic ordering enhance generation quality.
\section{Conclusion}
\label{sec:conclusion}

In summary, we introduce a novel autoregressive model that significantly enhances the generation of semantically enriched 3D house wireframes. 
Leveraging a unified, wire-based representation and semantic sequence reordering, our approach outperforms existing methods in accuracy, novelty, and semantic integrity. 
Our experiments on a diverse 3D house dataset underline our model's superior capability, positioning it as a valuable asset for applications in 3D modeling, computer-aided design, and virtual reality.

In the future, we aim to explore the deeper integration of semantic information related to house and architectural wireframes through various conditional generation methods, including controllable text, images, and point clouds.

\section*{Acknowledgements}
We thank all the anonymous reviewers for their insightful comments. 
This work was supported in parts by NSFC (U21B2023, U2001206, 62161146005), Guangdong Basic and Applied Basic Research Foundation (2023B1515120026), DEGP Innovation Team (2022KCXTD025), Shenzhen Science and Technology Program (KQTD20210811090044003, RCJC20200714114435012, JCYJ20210324120213036), Guangdong Laboratory of Artificial Intelligence and Digital Economy (SZ), and Scientific Development Funds from Shenzhen University.
%
%
\bibliographystyle{splncs04}
\bibliography{Wire_ref}

\begin{thebibliography}{10}
\providecommand{\url}[1]{\texttt{#1}}
\providecommand{\urlprefix}{URL }
\providecommand{\doi}[1]{https://doi.org/#1}

\bibitem{aichholzer1996straight}
Aichholzer, O., Aurenhammer, F.: Straight skeletons for general polygonal figures in the plane. In: Comput. Combin.: Annu. Int. Conf. pp. 117--126 (1996)

\bibitem{aichholzer1996novel}
Aichholzer, O., Aurenhammer, F., Alberts, D., G{\"a}rtner, B.: A novel type of skeleton for polygons. Springer (1996)

\bibitem{beltagy2020longformer}
Beltagy, I., Peters, M.E., Cohan, A.: Longformer: The long-document transformer (2020)

\bibitem{Cao23WireframeNet}
Cao, L., Xu, Y., Guo, J., Liu, X.: Wireframenet: {A} novel method for wireframe generation from point cloud. Comput. Graph.  \textbf{115},  226--235 (2023)

\bibitem{qi2020Intelligent}
Chen, Q., Wu, Q., Tang, R., Wang, Y., Wang, S., Tan, M.: Intelligent home 3d: Automatic 3d-house design from linguistic descriptions only. In: IEEE Conf. Comput. Vis. Pattern Recog. pp. 12622--12631 (2020)

\bibitem{cheng2022autoregressive}
Cheng, A.C., Li, X., Liu, S., Sun, M., Yang, M.H.: Autoregressive 3d shape generation via canonical mapping. In: Eur. Conf. Comput. Vis. pp. 89--104. Springer (2022)

\bibitem{Ziya2023ICCV}
Erko{\c{c}}, Z., Ma, F., Shan, Q., Nie{\ss}ner, M., Dai, A.: Hyperdiffusion: Generating implicit neural fields with weight-space diffusion. In: Int. Conf. Comput. Vis. pp. 14254--14264 (2023)

\bibitem{esser2021taming}
Esser, P., Rombach, R., Ommer, B.: Taming transformers for high-resolution image synthesis. In: IEEE Conf. Comput. Vis. Pattern Recog. pp. 12873--12883 (2021)

\bibitem{Hamilton17SAGEConv}
Hamilton, W.L., Ying, Z., Leskovec, J.: Inductive representation learning on large graphs. In: Adv. Neural Inform. Process. Syst. pp. 1024--1034 (2017)

\bibitem{Kaiming16ResNet}
He, K., Zhang, X., Ren, S., Sun, J.: Deep residual learning for image recognition. In: IEEE Conf. Comput. Vis. Pattern Recog. pp. 770--778 (2016)

\bibitem{Jingyu2022Wavelet}
Hu, J., Hui, K.H., Liu, Z., Li, R., Fu, C.W.: Neural wavelet-domain diffusion for 3d shape generation, inversion, and manipulation. ACM Trans. Graph.  \textbf{43}(2),  16:1--16:18 (2024)

\bibitem{Ruizhen20Graph2Plan}
Hu, R., Huang, Z., Tang, Y., van Kaick, O., Zhang, H., Huang, H.: Graph2plan: learning floorplan generation from layout graphs. ACM Trans. Graph.  \textbf{39}(4),  118:1--118:14 (2020)

\bibitem{Kun18Learning}
Huang, K., Wang, Y., Zhou, Z., Ding, T., Gao, S., Ma, Y.: Learning to parse wireframes in images of man-made environments. In: IEEE Conf. Comput. Vis. Pattern Recog. pp. 626--635 (2018)

\bibitem{Ibing2023Octree}
Ibing, M., Kobsik, G., Kobbelt, L.: Octree transformer: Autoregressive 3d shape generation on hierarchically structured sequences. In: IEEE Conf. Comput. Vis. Pattern Recog. pp. 2698--2707 (2023)

\bibitem{solidgen23}
Jayaraman, P.K., Lambourne, J.G., Desai, N., Willis, K.D.D., Sanghi, A., Morris, N.J.W.: Solidgen: An autoregressive model for direct b-rep synthesis. Trans. Mach. Learn. Res.  \textbf{2023} (2023)

\bibitem{Koch_2019_CVPR}
Koch, S., Matveev, A., Jiang, Z., Williams, F., Artemov, A., Burnaev, E., Alexa, M., Zorin, D., Panozzo, D.: Abc: A big cad model dataset for geometric deep learning. In: IEEE Conf. Comput. Vis. Pattern Recog. (June 2019)

\bibitem{lee2022autoregressive}
Lee, D., Kim, C., Kim, S., Cho, M., Han, W.S.: Autoregressive image generation using residual quantization. In: IEEE Conf. Comput. Vis. Pattern Recog. pp. 11523--11532 (2022)

\bibitem{sicong2023tell2design}
Leng, S., Zhou, Y., Dupty, M.H., Lee, W.S., Joyce, S., Lu, W.: Tell2design: {A} dataset for language-guided floor plan generation. In: Rogers, A., Boyd{-}Graber, J.L., Okazaki, N. (eds.) Assoc. Comput. Linguist. pp. 14680--14697 (2023)

\bibitem{zewen2022CNN}
Li, Z., Liu, F., Yang, W., Peng, S., Zhou, J.: A survey of convolutional neural networks: Analysis, applications, and prospects. IEEE Trans. Neural Netw. Learn. Syst.  \textbf{33}(12),  6999--7019 (2022)

\bibitem{Yancong2020DeepHL}
Lin, Y., Pintea, S.L., van Gemert, J.C.: Deep hough-transform line priors. In: Eur. Conf. Comput. Vis. pp. 323--340 (2020)

\bibitem{Yujia21PC2WF}
Liu, Y., D'Aronco, S., Schindler, K., Wegner, J.D.: {PC2WF:} 3d wireframe reconstruction from raw point clouds. In: Int. Conf. Learn. Represent. (2021)

\bibitem{luo2021diffusion}
Luo, S., Hu, W.: Diffusion probabilistic models for 3d point cloud generation. In: IEEE Conf. Comput. Vis. Pattern Recog. pp. 2837--2845 (2021)

\bibitem{Yicheng22LC2WF}
Luo, Y., Ren, J., Zhe, X., Kang, D., Xu, Y., Wonka, P., Bao, L.: Learning to construct 3d building wireframes from 3d line clouds. In: Brit. Mach. Vis. Conf. p.~91 (2022)

\bibitem{Wenchao22How3D}
Ma, W., Tan, B., Xue, N., Wu, T., Zheng, X., Xia, G.: How-3d: Holistic 3d wireframe perception from a single image. In: Int. Conf. 3D Vision. pp. 596--605 (2022)

\bibitem{Matveev21Parametric}
Matveev, A., Artemov, A., Zorin, D., Burnaev, E.: 3d parametric wireframe extraction based on distance fields. In: Int. Conf. Artif. Intell. Pattern Recog. pp. 316--322 (2021)

\bibitem{mittal2022autosdf}
Mittal, P., Cheng, Y.C., Singh, M., Tulsiani, S.: Autosdf: Shape priors for 3d completion, reconstruction and generation. In: IEEE Conf. Comput. Vis. Pattern Recog. pp. 306--315 (2022)

\bibitem{nash2020polygen}
Nash, C., Ganin, Y., Eslami, S.A., Battaglia, P.: Polygen: An autoregressive generative model of 3d meshes. In: Int. Conf. Mach. Learn. pp. 7220--7229 (2020)

\bibitem{nauata2020housegan}
Nauata, N., Chang, K.H., Cheng, C.Y., Mori, G., Furukawa, Y.: House-gan: Relational generative adversarial networks for graph-constrained house layout generation. In: Eur. Conf. Comput. Vis. pp. 162--177 (2020)

\bibitem{nauata2021houseganpp}
Nauata, N., Hosseini, S., Chang, K.H., Chu, H., Cheng, C.Y., Furukawa, Y.: House-gan++: Generative adversarial layout refinement network towards intelligent computational agent for professional architects. In: IEEE Conf. Comput. Vis. Pattern Recog. pp. 13632--13641 (2021)

\bibitem{van2016conditional}
van~den Oord, A., Kalchbrenner, N., Espeholt, L., Kavukcuoglu, K., Vinyals, O., Graves, A.: Conditional image generation with pixelcnn decoders. In: Adv. Neural Inform. Process. Syst. pp. 4790--4798 (2016)

\bibitem{pathak2016context}
Pathak, D., Krahenbuhl, P., Donahue, J., Darrell, T., Efros, A.A.: Context encoders: Feature learning by inpainting. In: IEEE Conf. Comput. Vis. Pattern Recog. pp. 2536--2544 (2016)

\bibitem{Pautrat23DeepLSD}
Pautrat, R., Barath, D., Larsson, V., Oswald, M.R., Pollefeys, M.: Deeplsd: Line segment detection and refinement with deep image gradients. In: IEEE Conf. Comput. Vis. Pattern Recog. pp. 17327--17336 (2023)

\bibitem{razavi2019generating}
Razavi, A., van~den Oord, A., Vinyals, O.: Generating diverse high-fidelity images with {VQ-VAE-2}. In: Adv. Neural Inform. Process. Syst. pp. 14837--14847 (2019)

\bibitem{Aurko21Efficient}
Roy, A., Saffar, M., Vaswani, A., Grangier, D.: Efficient content-based sparse attention with routing transformers. Trans. Assoc. Comput. Linguistics  \textbf{9},  53--68 (2021)

\bibitem{shabani2022housediffusion}
Shabani, M.A., Hosseini, S., Furukawa, Y.: Housediffusion: Vector floorplan generation via a diffusion model with discrete and continuous denoising. In: IEEE Conf. Comput. Vis. Pattern Recog. pp. 5466--5475 (2023)

\bibitem{siddiqui2023meshgpt}
Siddiqui, Y., Alliegro, A., Artemov, A., Tommasi, T., Sirigatti, D., Rosov, V., Dai, A., Nießner, M.: Meshgpt: Generating triangle meshes with decoder-only transformers (2023)

\bibitem{Jiahui22WallPlan}
Sun, J., Wu, W., Liu, L., Min, W., Zhang, G., Zheng, L.: Wallplan: synthesizing floorplans by learning to generate wall graphs. ACM Trans. Graph.  \textbf{41}(4),  92:1--92:14 (2022)

\bibitem{sun2020pointgrow}
Sun, Y., Wang, Y., Liu, Z., Siegel, J., Sarma, S.: Pointgrow: Autoregressively learned point cloud generation with self-attention. In: Winter Conf. Appl. Comput. Vis. pp. 61--70 (2020)

\bibitem{Xuefeng22Coarse}
Tan, X., Zhang, D., Tian, L., Wu, Y., Chen, Y.: Coarse-to-fine pipeline for 3d wireframe reconstruction from point cloud. Comput. Graph.  \textbf{106},  288--298 (2022)

\bibitem{Upadhyay22FLNet}
Upadhyay, A., Dubey, A., Arora, V., Kuriakose, M.S., Agarawal, S.: Flnet: Graph constrained floor layout generation. In: Int. Conf. Multimedia Expo Workshops. pp.~1--6 (2022)

\bibitem{Vaswani17transformer}
Vaswani, A., Shazeer, N., Parmar, N., Uszkoreit, J., Jones, L., Gomez, A.N., Kaiser, L., Polosukhin, I.: Attention is all you need. In: Adv. Neural Inform. Process. Syst. pp. 5998--6008 (2017)

\bibitem{Vinyals2015Pointer}
Vinyals, O., Fortunato, M., Jaitly, N.: Pointer networks. In: Adv. Neural Inform. Process. Syst. pp. 2692--2700 (2015)

\bibitem{Shidong2023ActFloor}
Wang, S., Zeng, W., Chen, X., Ye, Y., Qiao, Y., Fu, C.: Actfloor-gan: Activity-guided adversarial networks for human-centric floorplan design. IEEE Trans. Vis. Comput. Graph.  \textbf{29}(3),  1610--1624 (2023)

\bibitem{Wenming19PlanGen}
Wu, W., Fu, X., Tang, R., Wang, Y., Qi, Y., Liu, L.: Data-driven interior plan generation for residential buildings. ACM Trans. Graph.  \textbf{38}(6),  234:1--234:12 (2019)

\bibitem{Xue2019AFM}
Xue, N., Bai, S., Wang, F., Xia, G.S., Wu, T., Zhang, L.: Learning attraction field representation for robust line segment detection. In: IEEE Conf. Comput. Vis. Pattern Recog. pp. 1595--1603 (2019)

\bibitem{Xue23NEAT}
Xue, N., Tan, B., Xiao, Y., Dong, L., Xia, G.S., Wu, T.: Volumetric wireframe parsing from neural attraction fields (2023)

\bibitem{Xue23Holistically}
Xue, N., Wu, T., Bai, S., Wang, F., Xia, G., Zhang, L., Torr, P.H.S.: Holistically-attracted wireframe parsing: From supervised to self-supervised learning. IEEE Trans. Pattern Anal. Mach. Intell.  \textbf{45}(12),  14727--14744 (2023)

\bibitem{yan2022shapeformer}
Yan, X., Lin, L., Mitra, N.J., Lischinski, D., Cohen-Or, D., Huang, H.: Shapeformer: Transformer-based shape completion via sparse representation. In: IEEE Conf. Comput. Vis. Pattern Recog. pp. 6239--6249 (2022)

\bibitem{yu2023language}
Yu, L., Lezama, J., Gundavarapu, N.B., Versari, L., Sohn, K., Minnen, D., Cheng, Y., Gupta, A., Gu, X., Hauptmann, A.G., Gong, B., Yang, M.H., Essa, I., Ross, D.A., Jiang, L.: Language model beats diffusion -- tokenizer is key to visual generation (2023)

\bibitem{yue2023connecting}
Yue, Y., Kontogianni, T., Schindler, K., Engelmann, F.: {Connecting the Dots: Floorplan Reconstruction Using Two-Level Queries}. In: IEEE Conf. Comput. Vis. Pattern Recog. (2023)

\bibitem{zeng2022lion}
Zeng, X., Vahdat, A., Williams, F., Gojcic, Z., Litany, O., Fidler, S., Kreis, K.: {LION:} latent point diffusion models for 3d shape generation. In: Adv. Neural Inform. Process. Syst. (2022)

\bibitem{zhang2019graph}
Zhang, S., Tong, H., Xu, J., Maciejewski, R.: Graph convolutional networks: a comprehensive review. Comput. Soc. Netw.  \textbf{6}(1),  1--23 (2019)

\bibitem{Zhou2021PVD}
Zhou, L., Du, Y., Wu, J.: 3d shape generation and completion through point-voxel diffusion. In: Int. Conf. Comput. Vis. pp. 5826--5835 (2021)

\bibitem{Yichao19EndtoEnd}
Zhou, Y., Qi, H., Ma, Y.: End-to-end wireframe parsing. In: Int. Conf. Comput. Vis. pp. 962--971 (2019)

\bibitem{Yichao19Manhattan}
Zhou, Y., Qi, H., Zhai, Y., Sun, Q., Chen, Z., Wei, L., Ma, Y.: Learning to reconstruct 3d manhattan wireframes from a single image. In: Int. Conf. Comput. Vis. pp. 7697--7706 (2019)

\end{thebibliography}


\begin{thebibliography}{10}
\providecommand{\url}[1]{\texttt{#1}}
\providecommand{\urlprefix}{URL }
\providecommand{\doi}[1]{https://doi.org/#1}

\bibitem{aichholzer1996novel}
Aichholzer, O., Aurenhammer, F., Alberts, D., G{\"a}rtner, B.: A novel type of skeleton for polygons. Springer (1996)

\bibitem{beltagy2020longformer}
Beltagy, I., Peters, M.E., Cohan, A.: Longformer: The long-document transformer (2020)

\bibitem{Hamilton17SAGEConv}
Hamilton, W.L., Ying, Z., Leskovec, J.: Inductive representation learning on large graphs. In: Adv. Neural Inform. Process. Syst. pp. 1024--1034 (2017)

\bibitem{Kaiming16ResNet}
He, K., Zhang, X., Ren, S., Sun, J.: Deep residual learning for image recognition. In: IEEE Conf. Comput. Vis. Pattern Recog. pp. 770--778 (2016)

\bibitem{lee2022autoregressive}
Lee, D., Kim, C., Kim, S., Cho, M., Han, W.S.: Autoregressive image generation using residual quantization. In: IEEE Conf. Comput. Vis. Pattern Recog. pp. 11523--11532 (2022)

\bibitem{luo2021diffusion}
Luo, S., Hu, W.: Diffusion probabilistic models for 3d point cloud generation. In: IEEE Conf. Comput. Vis. Pattern Recog. pp. 2837--2845 (2021)

\bibitem{nash2020polygen}
Nash, C., Ganin, Y., Eslami, S.A., Battaglia, P.: Polygen: An autoregressive generative model of 3d meshes. In: Int. Conf. Mach. Learn. pp. 7220--7229 (2020)

\bibitem{Aurko21Efficient}
Roy, A., Saffar, M., Vaswani, A., Grangier, D.: Efficient content-based sparse attention with routing transformers. Trans. Assoc. Comput. Linguistics  \textbf{9},  53--68 (2021)

\bibitem{siddiqui2023meshgpt}
Siddiqui, Y., Alliegro, A., Artemov, A., Tommasi, T., Sirigatti, D., Rosov, V., Dai, A., Nießner, M.: Meshgpt: Generating triangle meshes with decoder-only transformers (2023)

\bibitem{Wenming19PlanGen}
Wu, W., Fu, X., Tang, R., Wang, Y., Qi, Y., Liu, L.: Data-driven interior plan generation for residential buildings. ACM Trans. Graph.  \textbf{38}(6),  234:1--234:12 (2019)

\bibitem{yu2023language}
Yu, L., Lezama, J., Gundavarapu, N.B., Versari, L., Sohn, K., Minnen, D., Cheng, Y., Gupta, A., Gu, X., Hauptmann, A.G., Gong, B., Yang, M.H., Essa, I., Ross, D.A., Jiang, L.: Language model beats diffusion -- tokenizer is key to visual generation (2023)

\bibitem{zeng2022lion}
Zeng, X., Vahdat, A., Williams, F., Gojcic, Z., Litany, O., Fidler, S., Kreis, K.: {LION:} latent point diffusion models for 3d shape generation. In: Adv. Neural Inform. Process. Syst. (2022)

\bibitem{Zhou2021PVD}
Zhou, L., Du, Y., Wu, J.: 3d shape generation and completion through point-voxel diffusion. In: Int. Conf. Comput. Vis. pp. 5826--5835 (2021)

\end{thebibliography}
\end{document}



\title{Generating 3D House Wireframes with Semantics} 
\subtitle{(Supplementary Materials)}

\titlerunning{3D House WireFrames}

\author{Xueqi Ma\orcidlink{0009-0004-0203-8501} \and
Yilin Liu\orcidlink{0000-0001-7336-1956} \and
Wenjun Zhou\orcidlink{0000-0003-1790-4201} \and
Ruowei Wang\orcidlink{0009-0003-9112-1712} \and
Hui Huang\thanks{Corresponding author}\orcidlink{0000-0003-3212-0544}}

\authorrunning{X.~Ma, Y.~Liu, W.~Zhou, R.~Wang, and H.~Huang}

\institute{Visual Computing Research Center, Shenzhen University \\
\email{hhzhiyan@gmail.com}\\
}

\maketitle

\section{Data}
\label{sec:data}

\begin{figure}[t!]
    \centering
    \includegraphics[width=0.90\textwidth]{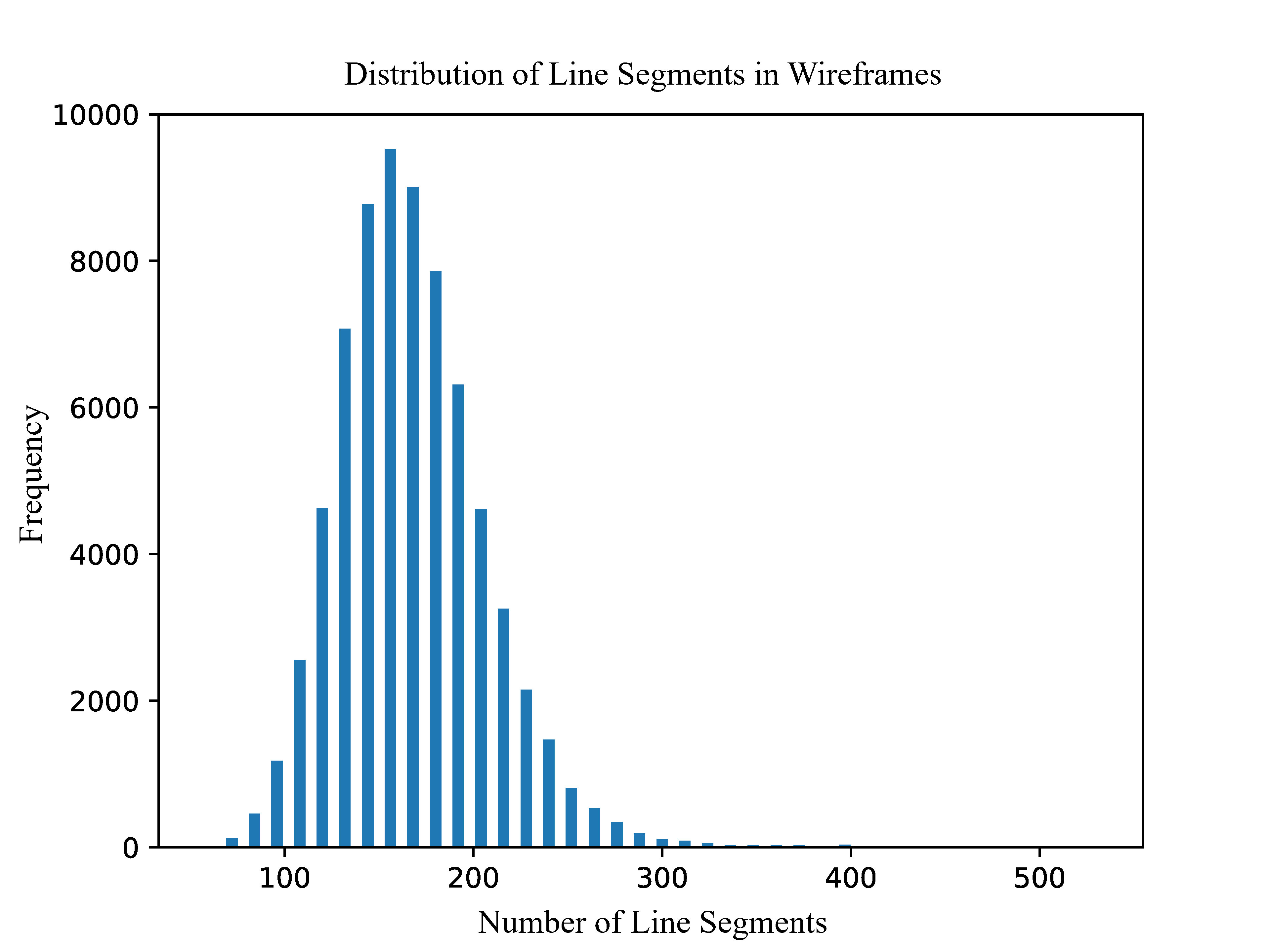}
    \caption{
    The distribution of the number of line segments in wireframes within the 3D house wireframe dataset. The horizontal axis represents the count of line segments, and the vertical axis shows the number of wireframe samples with that count. 
    }
    \label{fig_supp:dist}
\end{figure}

\paragraph{3D House Wireframe Dataset:}
Existing datasets for house models primarily feature 2D floor plans, which do not suffice for creating 3D house wireframes. 
Recognizing this deficiency, we have developed a comprehensive 3D house wireframe dataset. 
A typical 3D wireframe of a house encompasses 3 main components: the roof, exterior walls, and interior rooms. 
Our methodology initiated with extracting the 2D layout for critical room and exterior wall segments from the RPLAN dataset\cite{Wenming19PlanGen}. 
Subsequent steps involved lifting the corner points of these segments to establish the groundwork for the house's basic wireframe structure. 
To complete the wireframe model, we utilized the straight skeleton algorithm\cite{aichholzer1996novel} on the exterior walls, facilitating the construction of the roof's wireframe. 
Our analysis focused on wireframes containing fewer than 400 line segments, as depicted in \cref{fig_supp:dist}. 
This selection process resulted in the accumulation of 78,791 wireframes. 
We allocated these wireframes into training and test sets following a 9:1 ratio and normalized each wireframe to ensure its central alignment at the origin within a cubic space measured as \( [-1,1]^3 \).
For more information on downloading and using the dataset, please visit our GitHub repository: 
\href{https://github.com/3d-house-wireframe/3d-house-wireframe-dataset}{https://github.com/3d-house-wireframe/3d-house-wireframe-dataset}.

\begin{figure}[t!]
    \centering
    \includegraphics[width=0.5\textwidth]{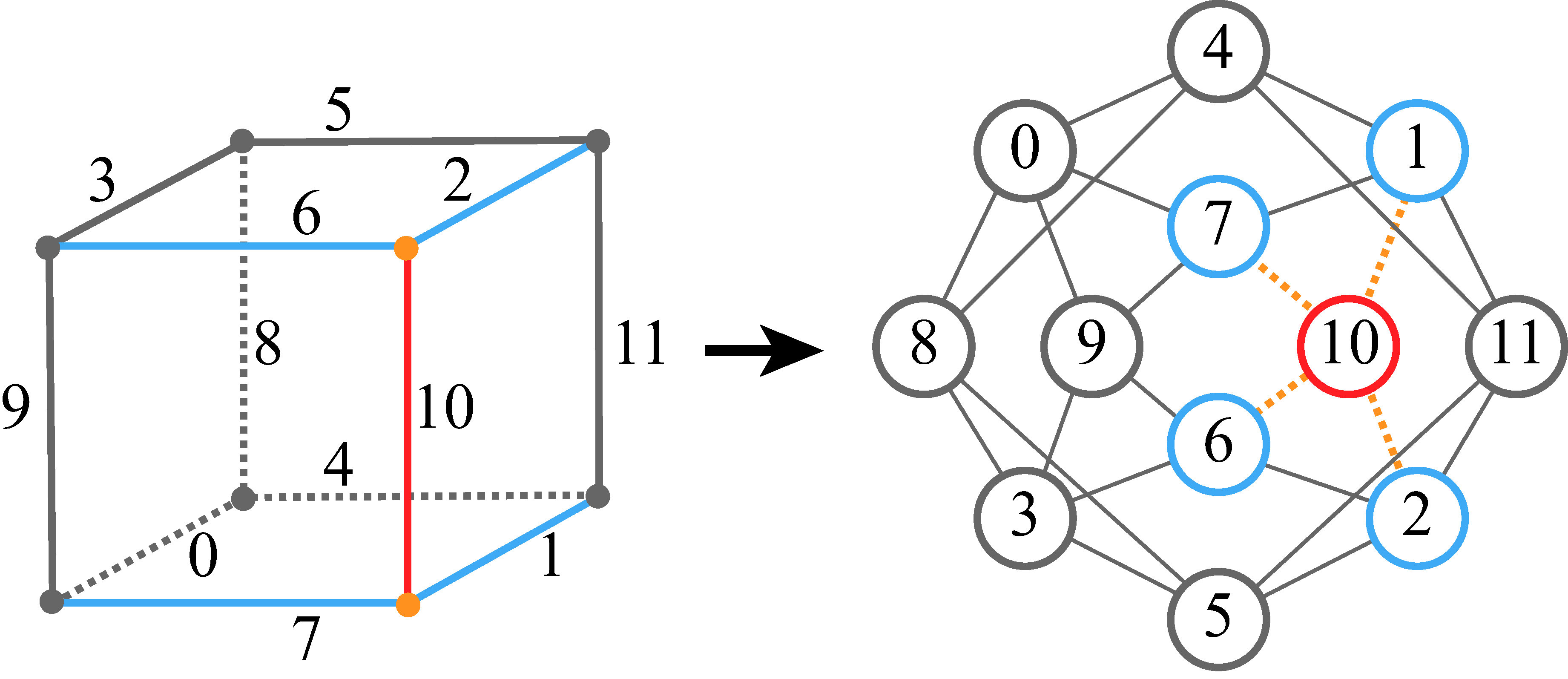}
    \caption{
    The process of transforming a wireframe into a graph.
    }
    \label{fig_supp:wf2graph}
\end{figure}
\begin{figure}[t!]
    \centering
    \includegraphics[width=0.80\textwidth]{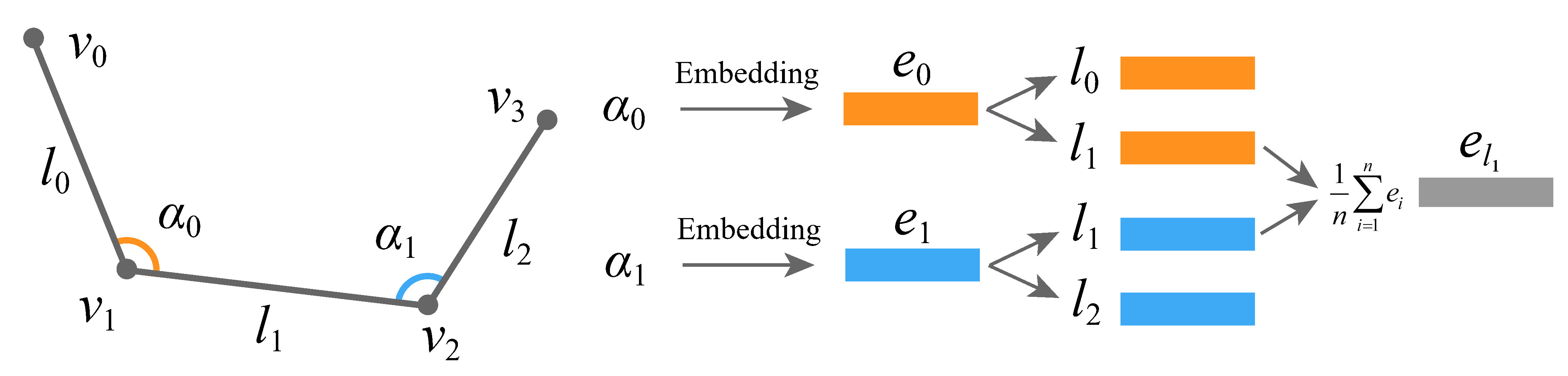}
    \caption{
    Embedding of angular features between adjacent line segments.
    }
    \label{fig_supp:node_angle}
\end{figure}

\paragraph{Data augmentation:}
Throughout the training phase, we applied various data augmentation techniques to improve the model's generalization ability.
Our augmentation strategies included:
1) Rotating the wireframes at predetermined angles, specifically \(\{0^\circ, 90^\circ, 180^\circ, 270^\circ\}\), to simulate different orientations;
2) Applying mirror flips across the YOZ plane to reflect the wireframes, thereby increasing the diversity of the training data; 
3) Adjusting the scale and position of the wireframe vertices in the x, y, and z dimensions. The scaling adjustments were confined to a range of \([0.9, 1.1]\), and translations were applied within a range of \([-0.1, 0.1]\). These transformations were intended to simulate variations in size and spatial alignment, further contributing to the robustness of the training.

\section{Method Details}
\subsection{Autoencoder}

\paragraph{Graph construction for feature learning:}
To facilitate effective feature learning of individual line segments, we convert the 3D wireframe models into graph representations, where nodes represent line segments and edges represent junctions.
As illustrated in \cref{fig_supp:wf2graph}, the left part of the figure presents the cube's wireframe, indicating each line segment with specific indices. 
Conversely, the right part displays the generated graph. 
Here, a particular line segment from the wireframe, such as segment 10 depicted in red, aligns with node 10 in the graph. 
The segments that connect to it, identified by indices 1, 2, 6, and 7 in the wireframe, are mirrored as connected nodes in blue within the graph. 
Furthermore, the graph visualizes the junctions, shown in yellow on the wireframe, as edges in the graph, effectively mapping the structural connections between line segments.

\paragraph{Features:}
Our encoder consists of a graph convolution module and a Local Multi-Head Attention module.
We introduced multiple features for each node in the graph (representing a line segment).
These features include 6 coordinates of the line segment, its length, its direction, and the 3 coordinates of the segment's midpoint.
To further enhance the node features, we calculate the angles between each pair of adjacent line segments and integrate these angular features into the corresponding line segment features, as shown in \cref{fig_supp:node_angle}.

\paragraph{Network:}
In constructing our neural network model, we first introduced a 5-layer Graph Convolutional Network (GCN) \cite{Hamilton17SAGEConv} as the initial encoder \(E_G\).
The feature dimensions of these layers are \([64, 128, 256, 256, 384]\). 
Subsequently, we added a 4-layer Local Multi-Head Attention (LMH Attention) module \(E_A\) \cite{Aurko21Efficient, beltagy2020longformer} to the encoder, with each layer having a dimension of 384. 
In the decoder, we first employed a 2-layer LMH Attention module \(D_A\) with each layer dimensioned at 384, followed by a 1D ResNet34 \cite{Kaiming16ResNet} as the second module \(D_R\). 
\(D_R\) comprises four sets of residual blocks, with the number of blocks in each set being \([3, 4, 6, 3]\), and the feature dimensions sequentially are \([128, 192, 256, 384]\).

Our decoder is designed to map the features of line segments into a \(128^3\) cubic space, facilitating the generation of discretized line segments. 
The output comprises the logits of 6 discrete coordinates for each line segment in this cubic space. 
For the LMH Attention, we use a window size of 64 and a dimension of 32 for each head. 
Additionally, the size of our codebook is set at 8192.

For the transformer model, we implemented a phased strategy that progresses from coarse to fine \cite{lee2022autoregressive} for autoregressively predicting the indices in the codebook. 
In this process, the transformer at the coarse stage is configured with 12 layers, a feature dimension of 512, and 8 heads. 
Following this, the transformer at the fine stage consists of 2 layers with a feature dimension of 512 and 8 heads. 
Both MeshGPT and our model have a maximum sequence length of 1624, whereas PolyGen's maximum is 816 for both vertices and segments due to its two-stage generation process. The temperature of all methods is set to 1.0, and the generation process stops when the termination symbol is predicted or the maximum sequence length is reached.

\subsection{Residual LFQ}

As shown in MeshGPT~\cite{siddiqui2023meshgpt}, the quantized feature is crucial for the transformer model to predict high-quality 3D models.
We used Residual LFQ \cite{lee2022autoregressive, yu2023language} to quantize the vertex features of the line segments.
Residual LFQ quantizes vertex features by treating them as Cartesian products of single-dimensional variables. 
Specifically, for a vertex feature vector \( z \), its quantized representation \( f(z) \) is defined as the value closest to each dimension of \( z \) in the codebook \( C_i \). 
Since each \( C_i \) only contains two values, -1 and 1, the quantized result for each dimension \( f(z_i) \) can be directly determined by the sign of \( z_i \):
\begin{equation}
    f(z_i) = \textbf{sign}(z_i) =
    \begin{cases} 
    -1 & \text{if } z_i \leq 0 \\
    1 & \text{if } z_i > 0,
    \end{cases}
\end{equation}
where \( z_i \) is the \( i \)th dimension of \( z \).

LFQ eliminates the need for the codebook lookup step typically required in traditional quantization, as each dimension's quantization index is obtained simply by \( f(z_i) = \text{sign}(z_i) \).
The token index for \( f(z) \) is then calculated by
\begin{equation}
    \text{Index}(z) = \sum_{i=1}^{n} 2^{i-1} \cdot \mathbb{I}\{z_i > 0\},  n = \mathrm{log}_2 K,
\end{equation}
where \( K \) is the codebook size, and \( \mathbb{I}\{z_i > 0\} \) is the indicator function, which equals 1 if \( z_i > 0 \), and 0 otherwise.

We adopt commit loss \cite{lee2022autoregressive} to impose constraints on the quantization process.
Additionally, to enhance the utilization of the codebook, we employ an entropy penalty \cite{yu2023language}. This not only aids the network in making more confident predictions but also encourages using more codes from the codebook.

\subsection{Loss Function for Autoencoder}

In our method, a line segment consists of two vertices, A and B, with each vertex's coordinate predicted from a discrete set of possible values ranging from 0 to 127. 
We use the cross-entropy loss function to optimize the autoencoder, which measures the discrepancy between the predicted probabilities and the ground truth discrete coordinates.

The predicted coordinate is represented as a probability distribution across the possible coordinate values for a given vertex on a line segment. 
The probability that the model predicts the coordinate \(c\) of vertex \( j \) (with \(j=1\) for vertex \(A\) and \(j=2\) for vertex \(B\)) of line segment \(i\) to be a particular value $k$ is denoted by \(p_{i,j,c,k}\). 
We uses smoothed one-hot encoding for true vertex coordinates, reducing penalties for physically closer coordinates. 
The true coordinate for this vertex is represented by \(y_{i,j,c,k}\).

The cross-entropy loss for each vertex is calculated using the formula:
\[ 
    L_{i,j} = - \frac{1}{3} \sum_{c=1}^{3} \sum_{k=0}^{127} y_{i,j,c,k}  \cdot \log(p_{i,j,c,k}), 
\]

Since each line segment has two vertices, the loss for line segment \(i\) is calculated as the average of the losses for both vertices \(A\) and \(B\), $ L_{i} = \tfrac{1}{2} (L_{i,1} + L_{i,2}) $.
The overall reconstruction loss for the model is the average of the total losses for each segment:
\[ 
    L_\text{recon} = \frac{1}{N} \sum_{i=1}^{N} L_{i}, 
\]
where \(N\) is the total number of line segments in the wireframe and $L_\text{recon}$ is the reconstruction loss function for the autoencoder.

\subsection{Loss Function for Transformer}
Our method uses a 2-layer residual quantization, representing each vertex by two tokens corresponding to indices in the codebook. 
Each line segment comprises two vertices, therefore being represented by 4 tokens.
With a total of \( N \) line segments, this equates to \( 4N \) tokens.
Our codebook size, \(|C|\), is 8192, allowing each token to have \( |C| \) possible values. 

The model predicts a probability for every possible value of each token. 
The probability that the model assigns to the \( j \)th token (where \( j \in \{1, 2, 3, 4\} \)) of the \( i \)th line segment being the \( c \)th token in the vocabulary is denoted as \( p_{i,j,c} \). 
The ground truth token is represented by  \( y_{i,j} \). Hence, the overall loss function is defined as:

\[ L_t = - \frac{1}{4N} \sum_{i=1}^{N} \sum_{j=1}^{4} \sum_{c=1}^{|C|} \mathbb{I}\{y_{i,j} = c\} \cdot \log(p_{i,j,c}), \]
where \( \mathbb{I}\{y_{i,j} = c\} \)  is the indicator function, which equals 1 if the true token \( y_{i,j} \) is equal to \(c\), and 0 otherwise, $L_t$ is the loss function for the transformer.

\subsection{Baselines}
For PolyGen \cite{nash2020polygen}, we utilize the official TensorFlow implementation provided by the authors. 
Regarding MeshGPT \cite{siddiqui2023meshgpt}, we replicate it based on the detailed descriptions provided in the paper. 
Given that MeshGPT was initially designed to generate triangular meshes, and we aim to generate wireframes, we adapt the MeshGPT during the replication process to shift its focus from predicting triangular faces to predicting line segments. 
We employ the same dataset for training and testing purposes for all methods under study.

\section{Experiment Details}
\subsection{Metric Details}
Following previous works on 3D generative models~\cite{luo2021diffusion, zeng2022lion, Zhou2021PVD}, we adopted COV, MMD, and 1-NN as our evaluation metrics. 
COV stands for \textit{Coverage}, measuring the extent to which generated samples cover the real samples. 
MMD, or Maximum Mean Discrepancy, quantifies the difference between the generated and real samples. 
Lastly, 1-NN, meaning Nearest Neighbor, assesses the similarity between generated samples and their nearest real counterparts.

The definitions of these metrics are as follows:
\[
    \text{COV}(S_g, S_r) = \frac{|\{\arg\min_{y \in S_r} D(X, Y) | X \in S_g\}|}{|S_r|},
\]
\[
    \text{MMD}(S_g, S_r) = \frac{1}{|S_r|} \sum_{Y \in S_r} \min_{X \in S_g} D(X, Y),    
\]
\[
    1\text{-NN}(S_g, S_r) = \frac{\sum_{x \in S_g} \mathbb{I}[N_x \in S_g] + \sum_{y \in S_r} \mathbb{I}[N_y \in S_r]}{|S_g| + |S_r|},
\]
where \(S_g\) and \(S_r\) represent the generated and real samples, respectively, and \(D\) denotes the distance function.

We utilized Chamfer distance and Earth Mover's Distance (EMD) as metrics to measure the similarity of wireframes. Chamfer distance, a method for quantifying point cloud similarities through nearest point distances, and EMD, which assesses the minimal effort required to transform one point cloud into another, were both applied to evaluate the wireframe comparisons effectively.
All the metrics are computed on 8192 generated samples, each with 4096 sample points on their segments.
The generated samples are compared with normalized augmented data, including rotation and axis flip.

To evaluate the structural validity of our generated 3D wireframes, we analyzed the relationships between the vertices and the line segments. 
This analysis is based on the following assumptions: if a vertex is only connected to a single line segment, it may indicate that one end of the segment is not connected to any other segment, resulting in the segment floating in space; if a vertex is connected to two line segments, it could suggest that the segments are located within the interior of the model's edges; whereas a vertex connected to three or more line segments typically indicates a structurally plausible wireframe vertex. 
Based on this understanding, we designed two metrics for quantifying the analysis: the Two-Line-Connected Vertex Proportion (2L-CVP), which measures the proportion of vertices connected to at least two line segments, and the Three-Line-Connected Vertex Proportion (3L-CVP), which is the proportion of vertices connected to at least three line segments. 
These two metrics collectively aid in evaluating the wireframes' structural validity, ensuring the generated wireframes' accuracy and realism.

\subsection{User Study}
\begin{figure}[t!]
    \centering
    \includegraphics[width=0.95\textwidth]{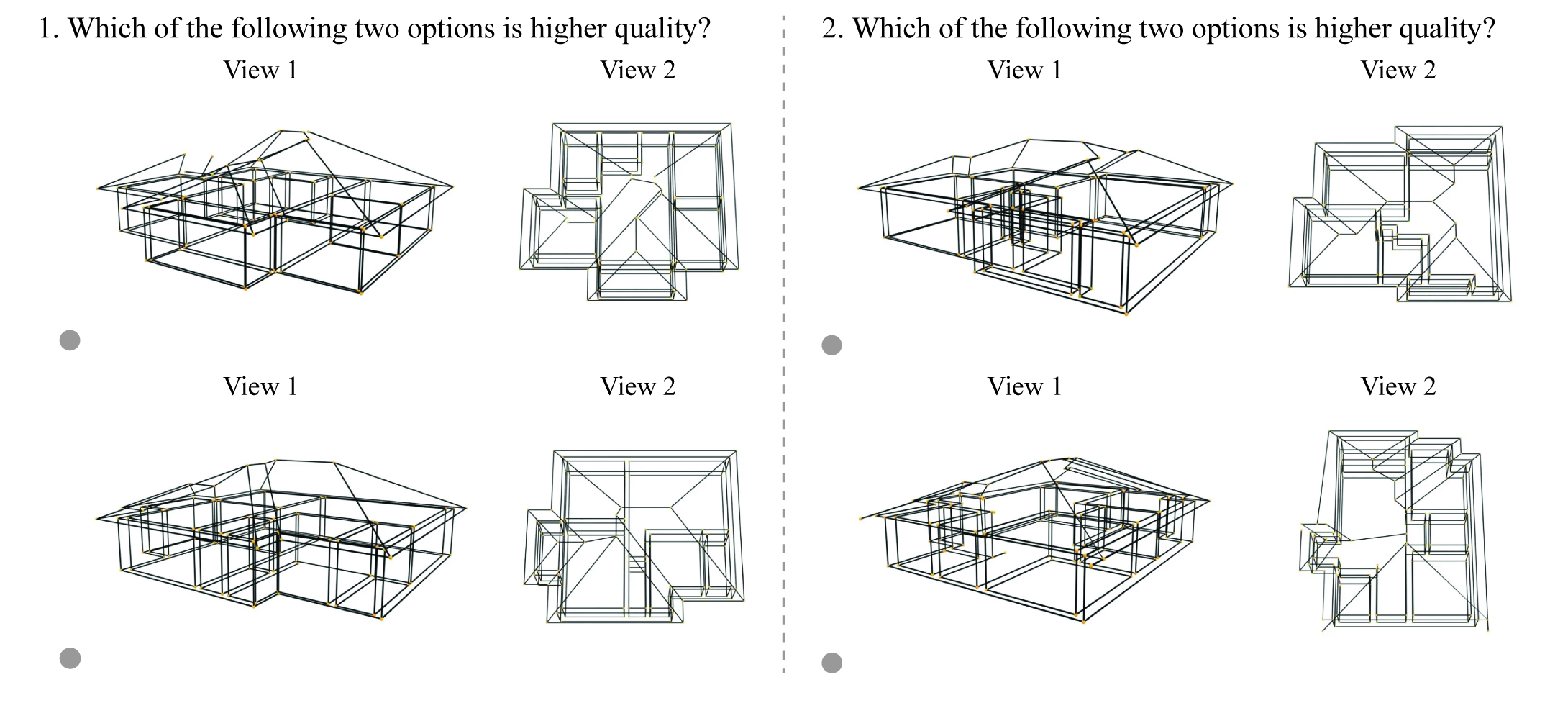}
    \captionsetup{skip=1pt}
    \caption{
    User study interface.
    }
    \label{fig_supp:user}
\end{figure}

As depicted in \cref{fig_supp:user}, we present the interface for our user study. Initially, we generated 1024 samples using various methods, and for comparison, we also randomly selected 1024 real wireframes from the dataset.

Since we have 4 methods (including ground truth), there are 6 possible pairings. For each pairing, we randomly selected 4 groups of samples, resulting in a total of 24 sample sets for our study.

These samples are shuffled before the presentation to ensure the display order does not influence the users' evaluations. Users could click on the images of each wireframe to view more details. We recorded the users' choices and calculated the win rates for each method. 60 participants were invited to participate in the survey, with each user evaluating 24 sets of samples. 
To assess the effects of different methods, we first calculated the proportion of user preferences between two methods. Taking our method and PolyGen \cite{nash2020polygen} as examples, 92\% of users preferred our method, while 8\% preferred PolyGen. Calculating the score difference revealed that our method outperformed PolyGen by 0.84 points, which is 0.84 = 0.92 (our selection rate) - 0.08 (PolyGen's selection rate).

\begin{figure}[t!]
    \centering
    \includegraphics[width=0.95\textwidth]{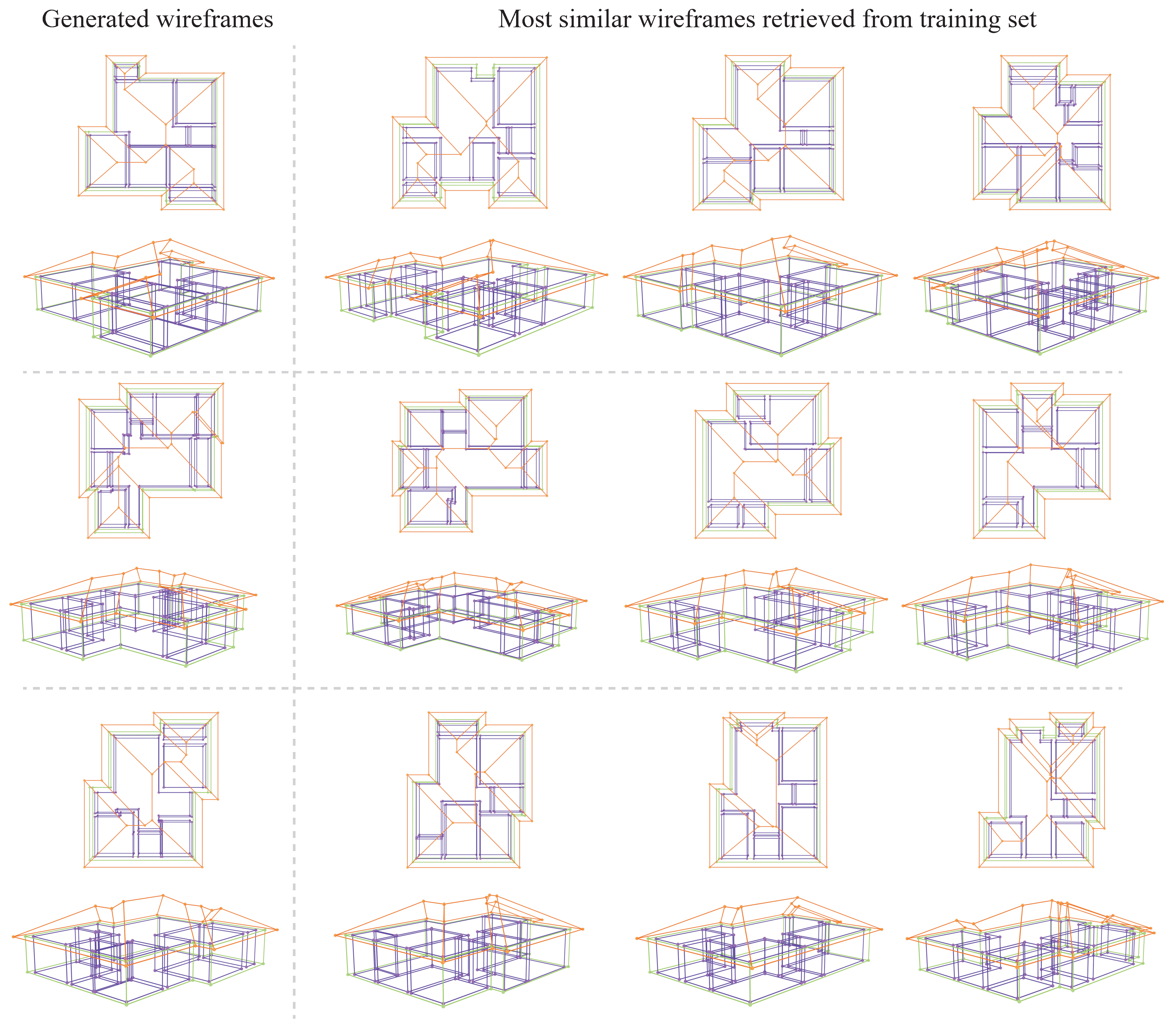}
    \captionsetup{skip=1pt}
    \caption{
        Novelty analysis of generated wireframes. We present a comparison of a wireframe produced by our method against its 3 nearest neighbors from the 3D house wireframe  training dataset, determined by Chamfer Distance (CD).
    }
    \label{fig_supp:top-3}
\end{figure}

\begin{figure}[t!]
    \centering
    \includegraphics[width=0.99\textwidth]{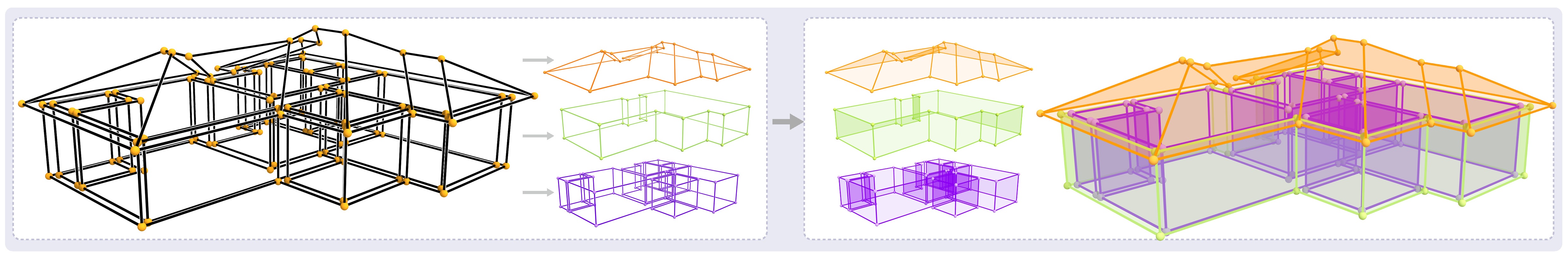}

    \caption{The reconstructed wireframe model can be easily split into several components. We also show their corresponding mesh on the right.}
    \label{fig_supp:three-comp}
\end{figure}

\begin{figure}[t!]
    \centering
    \includegraphics[width=0.99\textwidth]{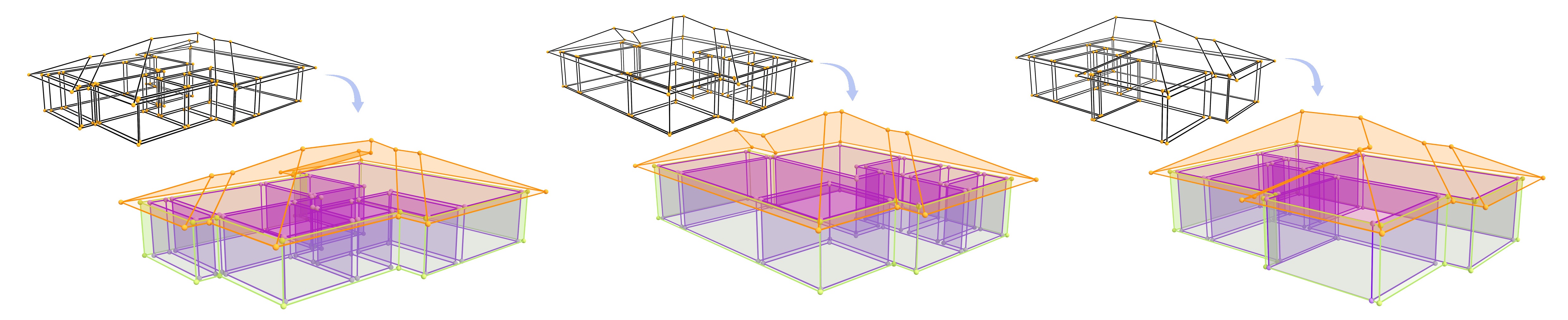}
    
    \caption{
    The resulting wireframe can be easily converted into a mesh model.
    }
    \label{fig_supp:recon}
\end{figure}

\subsection{Wireframe Novelty Analysis}
We conducted another novelty analysis on the wireframes generated by our method. As demonstrated in \cref{fig_supp:top-3}, we compare these generated wireframes with the 3 most similar real wireframes from our training dataset. Our findings reveal a significant difference between our generated wireframes and those from the dataset, indicating that our method can produce diverse wireframes.

\subsection{More Visual Results}

\begin{figure}[t]
    \centering
    \includegraphics[width=0.9\textwidth]{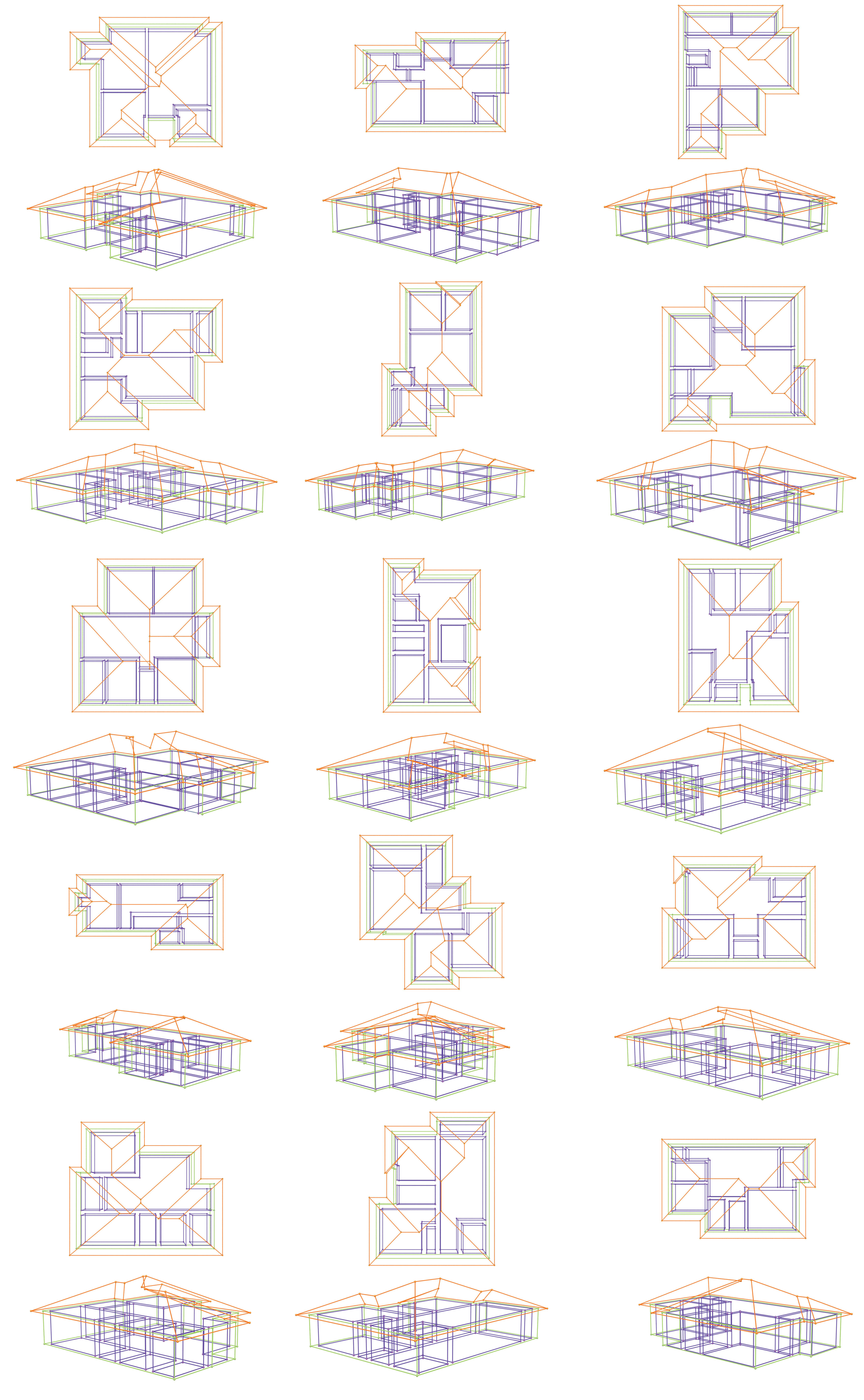}
    \caption{
        Unconditionally generated 3D house wireframes from our method.
    }
    \label{fig_supp:addit-1}
\end{figure}
\begin{figure}[t]
    \centering
    \includegraphics[width=0.9\textwidth]{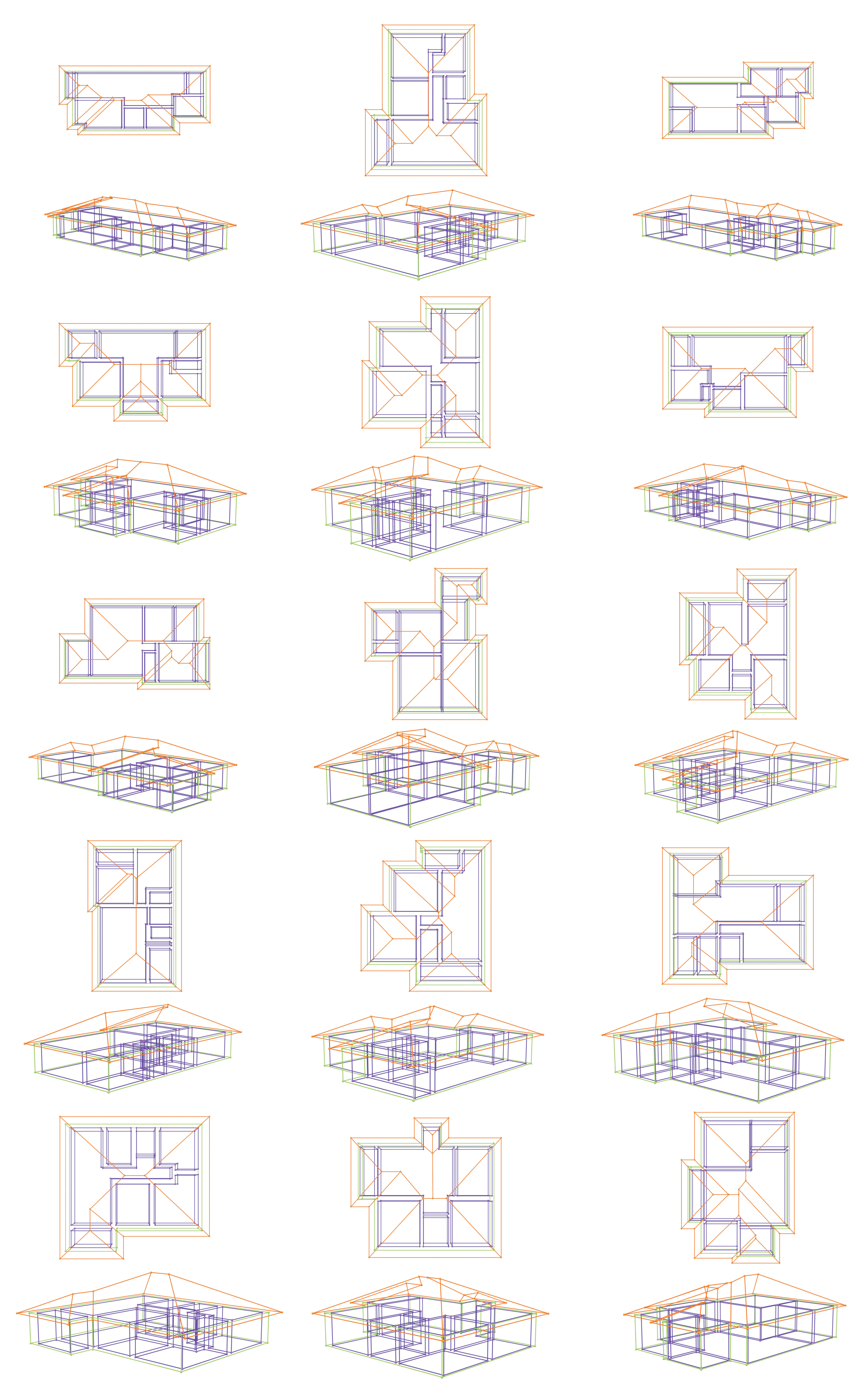}
    \caption{
        Unconditionally generated 3D house wireframes from our method.
    }
    \label{fig_supp:addit-2}
\end{figure}

\cref{fig_supp:three-comp} presents additional wireframe segmentation results. These wireframes are divided into multiple components, such as walls, roofs, and different rooms, based on the connectivity of the line segments. 
\cref{fig_supp:recon} shows more results of converting wireframes into mesh models. These results further demonstrate that our wireframes can be easily converted into mesh models. 
Additionally, we utilize our method to generate a variety of 3D house wireframes, as shown in \cref{fig_supp:addit-1} and \cref{fig_supp:addit-2}. We showcase the geometric characteristics of the wireframes and illustrate the diversity in different house layouts.

%
%

\section*{Acknowledgements}
We thank all the anonymous reviewers for their insightful comments. 
This work was supported in parts by NSFC (U21B2023, U2001206, 62161146005), Guangdong Basic and Applied Basic Research Foundation (2023B1515120026), DEGP Innovation Team (2022KCXTD025), Shenzhen Science and Technology Program (KQTD20210811090044003, RCJC20200714114435012, JCYJ20210324120213036), Guangdong Laboratory of Artificial Intelligence and Digital Economy (SZ), and Scientific Development Funds from Shenzhen University.

\bibliographystyle{splncs04}
\bibliography{Wire_ref}